\pdfoutput=1
\documentclass{article}

\usepackage[preprint]{neurips_2023}

\usepackage[utf8]{inputenc} 
\usepackage[T1]{fontenc}    
\usepackage{hyperref}       
\usepackage{url}            
\usepackage{booktabs}       
\usepackage{amsfonts}       
\usepackage{nicefrac}       
\usepackage{microtype}      
\usepackage[table,xcdraw]{xcolor}         

\usepackage{subcaption}
\usepackage{multicol}
\usepackage{graphicx}
\usepackage{wrapfig}
\usepackage{multirow}
\usepackage{makecell}
\usepackage{xspace}
\usepackage{amsmath}
\usepackage{listings}
\usepackage{algorithm}
\usepackage{amsfonts,amssymb}
\usepackage{mathrsfs}
\usepackage{subcaption}
\usepackage{natbib}
\setcitestyle{numbers,square}
\usepackage{graphicx}       
\usepackage{cleveref}
 \usepackage{minitoc}
\usepackage{color}
\usepackage{xcolor}
\usepackage{listings}

\usepackage[utf8]{inputenc}
\usepackage{CJKutf8}
\usepackage{arydshln} 
\usepackage{algorithm}
\usepackage{algorithmicx}
\usepackage{tikz}
\usepackage{asymptote}
\usepackage{tikz}
\usepackage{amsmath}
\usepackage{algorithm}
\usepackage{algpseudocode}
\usepackage{booktabs}
\usepackage{multirow}
\usepackage{microtype}
\usepackage{tabularx} 
\usepackage{longtable}
\usepackage{xspace}
\usepackage{enumitem}
\usepackage[normalem]{ulem}
\usepackage[para,online,flushleft]{threeparttable}
\useunder{\uline}{\ul}{}

\newcommand{\hide}[1]{} 
\newcommand{\vpara}[1]{\vspace{0.04in}\noindent\textbf{#1}\xspace}
\usepackage{multirow}

\newcommand{\hhide}[1]{}

\title{ChatGLM-Math: Improving Math Problem-Solving in Large Language Models with a Self-Critique Pipeline}

 \author{
Yifan Xu$^{12\dagger*}$, Xiao Liu$^{12*}$, Xinghan Liu$^{12\dagger}$, Zhenyu Hou$^{12}$, Yueyan Li$^{1\dagger}$, Xiaohan Zhang$^{1}$,\\
\bf{Zihan Wang$^{12}$, Aohan Zeng$^{12}$,Zhengxiao Du$^{12}$, Wenyi Zhao$^{1}$, Jie Tang$^{2}$, Yuxiao Dong$^{2}$}\\ \\
\textsuperscript{1}Zhipu.AI \quad
\textsuperscript{2}Tsinghua University
\\ \\
{\includegraphics[height=3.5ex]{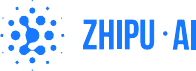}}}

\begin{document}

\maketitle
\renewcommand{\thefootnote}{\fnsymbol{footnote}}
    \footnotetext[1]{Yifan and Xiao contributed equally. Emails: \texttt{xu-yf23@mails.tsinghua.edu.cn},\texttt{shawliu9@gmail.com}}
    \footnotetext[2]{Work done while Xinghan and Yueyan interned at Zhipu AI.}
\renewcommand{\thefootnote}{\arabic{footnote}}

\vspace{-8mm}
\begin{abstract}
    Large language models (LLMs) have shown excellent mastering of human language, but still struggle in real-world applications that require mathematical problem-solving.
    While many strategies and datasets to enhance LLMs' mathematics are developed, it remains a challenge to simultaneously maintain and improve both language and mathematical capabilities in deployed LLM systems.
    In this work, we tailor the Self-Critique pipeline, which addresses the challenge in the feedback learning stage of LLM alignment.
    We first train a general Math-Critique model from the LLM itself to provide feedback signals. Then, we sequentially employ rejective fine-tuning and direct preference optimization over the LLM's own generations for data collection.
    Based on ChatGLM3-32B, we conduct a series of experiments on both academic and our newly created challenging dataset, \textsc{MathUserEval}. Results show that our pipeline significantly enhances the LLM's mathematical problem-solving while still improving its language ability, outperforming LLMs that could be two times larger.
    Related techniques have been deployed to ChatGLM\footnote{\url{https://chatglm.cn}}, an online serving LLM. Related evaluation dataset and scripts are released at \url{https://github.com/THUDM/ChatGLM-Math}.
\end{abstract}

\begin{figure}[h]
    \centering
    \begin{minipage}[b]{0.49\textwidth}
        \centering

        \renewcommand\tabcolsep{1.5pt}
        \renewcommand\arraystretch{1.2}
        \resizebox{\textwidth}{!}{%
        \begin{tabular}{@{}lcc@{}}

        \toprule
        Model & \makecell{Avg. of GSM8k \\\& MATH} & \makecell{AlignBench\\Language} \\
        \midrule
        \footnotesize DeepSeek-67B-Chat~\cite{deepseekai2024deepseek} & 58.3 & 7.11 \\
        \footnotesize DeepSeek-67B-Chat-DPO~\cite{deepseekai2024deepseek} & 57.7 (\textbf{\textcolor{red}{-1.2\%}}) & 7.60 (\textbf{\textcolor{teal}{+6.8\%}}) \\
        \midrule
        \footnotesize InternLM2-Chat-20B~\cite{2023internlm} & 57.2 & 7.68 \\
        \footnotesize Math-InternLM2-20B~\cite{2023internlm} & 60.2 (\textbf{\textcolor{teal}{+5.1\%}}) & 6.53 (\textbf{\textcolor{red}{-14.8\%}}) \\
        \midrule
        \footnotesize ChatGLM3-32B-SFT-2312 & 52.4 & 7.37 \\
        \footnotesize \qquad\qquad\qquad + RFT\&DPO & 61.6 (\textbf{\textcolor{teal}{+17.5\%}}) & 7.80 (\textbf{\textcolor{teal}{+5.85\%}}) \\
        \bottomrule
        \end{tabular}
        }
        \captionof{table}{Our self-critique pipeline enables simultaneous improvement of language and mathematical abilities. Previous alignment methods enhance language but could potentially impair mathematical abilities~\cite{deepseekai2024deepseek}, whereas math-specialized models could harm language capabilities~\cite{2023internlm}. }
        \label{tab:first table}
        
        \vspace{1mm}
    \end{minipage}
    \hfill
    \begin{minipage}[b]{0.5\textwidth}
        \centering
        \includegraphics[width=\textwidth]{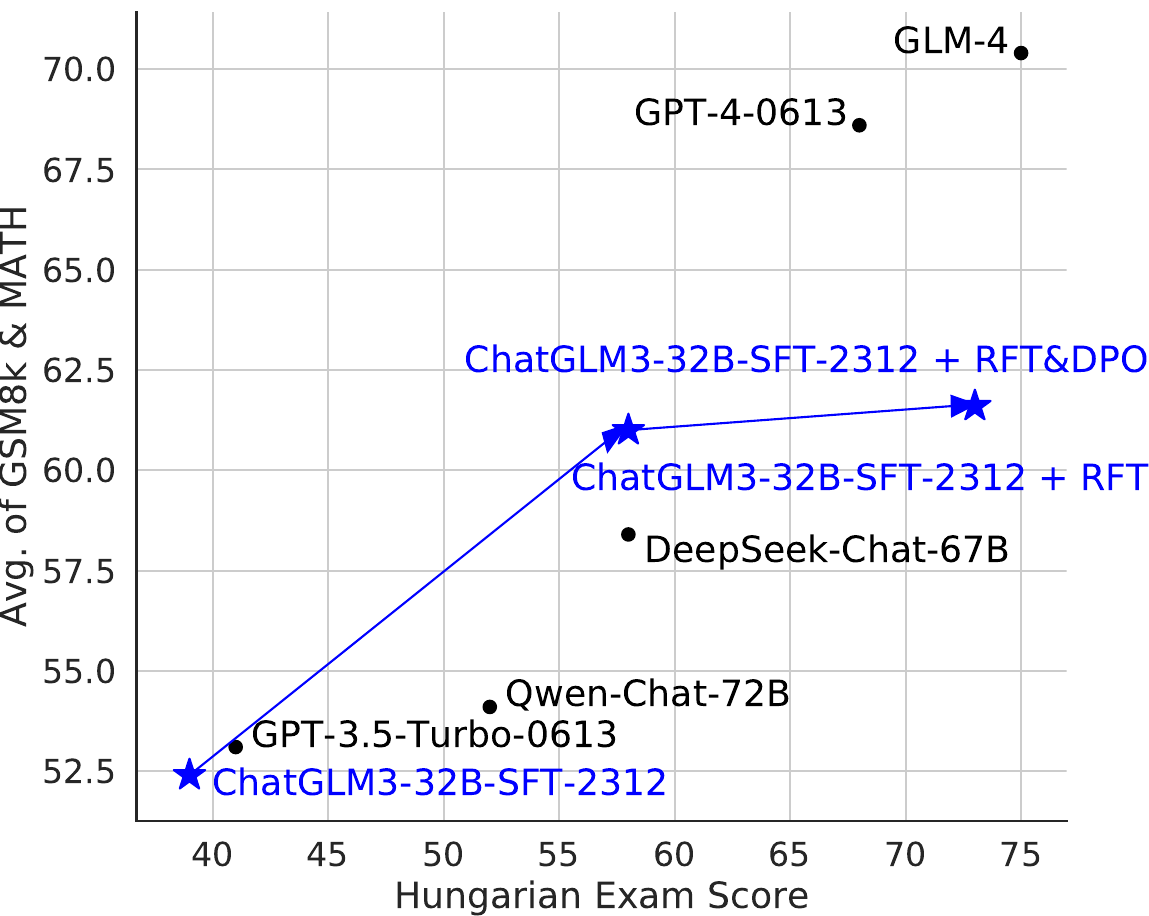}
        \caption{Results of Hungarian Exam and Average Scores of GSM8k and MATH.}
        \label{fig:hungarg test}
        \vspace{-5mm}
    \end{minipage}
\end{figure}

\section{Introduction}
Large Language Models (LLMs)~\cite{brown2020language, chowdhery2022palm, kaplan2020scaling, scao2022bloom, touvron2023llama, zeng2022glm, Claude} have garnered widespread attention for their remarkable proficiency in various linguistic tasks such as text summarization\cite{NIPS2015_afdec700, volske-etal-2017-tl, xsum-emnlp, li-etal-2022-csl}, question answering~\cite{hendrycks2020measuring, kwiatkowski2019natural, bras_Gao_Choi_2020}, and role-playing conversations~\cite{tu2024charactereval, zhou2023characterglm, shao2023characterllm}. 
Furthermore, their potential in addressing complex problems requiring mathematical reasoning~\cite{metamath,wang2023mathshepherd,wizardmath} has expanded their applicability across real-world missions~\cite{liu2023agentbench,bai2023longbench}.

Despite these advances, optimizing LLMs to excel simultaneously in language understanding and mathematical problem-solving presents a notable challenge. 
The prevalent reinforcement learning from human feedback (RLHF) approach primarily enhances text generation based on reward models reflecting human preferences~\cite{touvron2023llama, ouyang2022training, touvron2023llama2}. 
Although this method boosts the quality of generated text, it often overlooks the accuracy and logical coherence essential for solving mathematical problems, leading to a discrepancy in performance known as the "alignment tax"\cite{askell2021general} when applied to mathematical reasoning (refer to Table~\ref{tab:first table}). 
Conversely, attempts to bolster LLMs’ mathematical capabilities typically entail supervised fine-tuning (SFT) that inadvertently diminishes their linguistic versatility, posing a dilemma for practical applications of LLM systems~\cite{2023internlm,metamath,wizardmath,yue2023mammoth}.

\vpara{Pipeline: Self-Critique.}
This paper introduces a novel approach aimed at enhancing both linguistic and mathematical skills of LLMs without compromising one for the other. 
Our strategy deviates from traditional RLHF by incorporating a Math-Critique model derived from the LLM itself, which evaluates its mathematical outputs. 
This self-critique mechanism enables the model to learn from AI-generated feedback specifically tailored to mathematical content~\cite{bai2022constitutional, lee2023rlaif}. Our methodology comprises two primary phases:

\begin{itemize}[leftmargin=*,itemsep=0pt,parsep=0.2em,topsep=0.2em,partopsep=0.0em]
\item \textbf{Stage 1: Rejective Fine-tuning (RFT)}~\cite{yuan2023scaling-mathrft} employs a rejection sampling technique, wherein responses failing to meet Math-Critique standards are discarded, while the rest undergo further fine-tuning. This stage aims to enhance the model's accuracy and consistency in mathematical responses while ensuring diversity among the selected answers.
\item \textbf{Stage 2: Direct Preference Optimization (DPO)}~\cite{rafailov2023direct} extends the improvement process by directly learning from pairs of correct and incorrect answers, further refined through Math-Critique, focusing on the most challenging questions from the previous stage.
\end{itemize}

\vpara{Benchmark: \textsc{MathUserEval}.}
To accurately assess LLMs’ capabilities in solving real-world mathematical problems, we develop the \textsc{MathUserEval} dataset. 
It features a diverse range of questions, extending beyond academic exercises to include practical application scenarios, thereby better reflecting user needs compared to traditional academic math datasets~\cite{zhao2020ape210k,wang-etal-2017-deep-math23,cobbe2021training}. 
We leverage both GPT-4-turbo and our Math-Critique model for comprehensive scoring.

In summary, our contributions include:

\begin{itemize}[leftmargin=*,itemsep=0pt,parsep=0.2em,topsep=0.2em,partopsep=0.0em]
\item The introduction of the Self-Critique pipeline, a novel framework that elevates both the mathematical and linguistic capabilities of LLMs through self-generated feedback, thereby eliminating the need for external supervisory models and manual annotations. This approach has been validated on a ChatGLM3-32B model, achieving unparalleled performance on the \textsc{MathUserEval}, Ape210k~\cite{zhao2020ape210k}, MATH~\cite{hendrycks2020measuring}, and the linguistic tasks of AlignBench~\cite{liu2023alignbench}.

\item The creation of the \textsc{MathUserEval} benchmark, tailored to assess LLMs on complex, open-ended mathematical queries relevant to real-world applications, setting a new standard in evaluating practical mathematical reasoning capabilities.

\item A detailed analysis of the key factors contributing to the enhancement of mathematical proficiency through the Self-Critique pipeline, offering insights into future directions for autonomous model improvement.
\end{itemize}

\hide{
\section{Introduction}
Large language models (LLMs,~\cite{brown2020language, chowdhery2022palm, kaplan2020scaling, scao2022bloom, touvron2023llama, zeng2022glm, Claude}) have been widely deployed due to its superior ability on language tasks, such as text summarization~\cite{NIPS2015_afdec700, volske-etal-2017-tl, xsum-emnlp, li-etal-2022-csl}, question answering~\cite{hendrycks2020measuring, kwiatkowski2019natural, bras_Gao_Choi_2020}, and role-play conversations~\cite{tu2024charactereval, zhou2023characterglm, shao2023characterllm}.
They have also demonstrated strong potential in solving advanced tasks that require mathematical reasoning~\cite{metamath,wang2023mathshepherd,wizardmath}, which broadens their prospect to be deployed in real-world applications~\cite{openai2023gpt4}.

However, in practical LLM deployment, it turns out nontrivial to simultaneously maintain and improve language and mathematical abilities.
On one hand, a critical challenge lies in reinforcement learning from the human feedback (RLHF) stage, where \textit{de facto} approaches reinforce LLMs against reward models (RMs) that reflect human preference~\cite{touvron2023llama, ouyang2022training, touvron2023llama2}.
Such a paradigm improves text generation quality, but it does not necessitate factual correctness and logical consistency crucial to mathematical problem-solving.
As a result, while many LLMs benefit from RLHF on language understanding, they suffer from the so-called ``alignment-tax''~\cite{askell2021general} on mathematical reasoning (Cf. Table~\ref{tab:first table}).
On the other hand, notwithstanding some recent attempts to exploit LLMs' mathematical ability~\cite{2023internlm,metamath,wizardmath,yue2023mammoth}, they usually address the supervised fine-tuning (SFT) stage and become math experts in the sacrifice of general language capability.
Thus, it is uneconomical and inconvenient for deployed LLM systems to adopt most of these strategies.

\vpara{Pipeline: Self-Critique.}
In this work, we endeavor to explore a ``win-win'' solution for both language and mathematical abilities for practical LLM deployment.
Instead of addressing the SFT stage, we highlight LLMs' mathematical enhancement in the feedback learning period during LLM alignment.
Different from RLHF which leverages human feedback, we employ a Math-Critique model derived from the target LLM itself to provide judging feedback on its generated solutions.
In other words, we reinforce LLMs from AI's feedback~\cite{bai2022constitutional, lee2023rlaif} on mathematics.
The pipeline encompasses two stages:

\begin{itemize}[leftmargin=*,itemsep=0pt,parsep=0.2em,topsep=0.2em,partopsep=0.0em]
    \item \textbf{Stage 1: Rejective Fine-tuning (RFT)}~\cite{yuan2023scaling-mathrft}  is a process where the model undergoes multiple sampling iterations, during which certain responses are rejected based on predefined criteria. In contrast, the remaining responses are utilized for supervised fine-tuning. We implement a rejection sampling method based on Math-Critique evaluations to refine the dataset, focusing on enhancing the LLM's ability to generate mathematically accurate and logically consistent responses. We also incorporate measures to ensure diversity and the selection of entirely correct responses. 
    \item \textbf{Stage 2: Direct Preference Optimization (DPO)} ~\cite{rafailov2023direct} updates the model directly by learning from a pair of correct-incorrect answers. Building on the foundation set by the RFT step, we continue to sample multiple times for problems that were challenging to resolve in the previous phase. We then use Math-Critique to re-rank these samples and select the pair of answers with the most significant difference for learning.
\end{itemize}

\vpara{Benchmark: \textsc{MathUserEval}.}
Additionally, to allow evaluation of LLMs on complicated real-world problems requiring mathematical reasoning, we design a corresponding \textsc{MathUserEval} dataset for testing.
Compared to existing academic math datasets~\cite{zhao2020ape210k,wang-etal-2017-deep-math23,hendrycks2020measuring,cobbe2021training}, it composes questions from the high school level and above, as well as real-world application problems to match users' needs. 
Many of these questions are challenging to evaluate with traditional methods (e.g., string matching), yet they are more representative of real-world use cases. 
We implement the evaluation on \textsc{MathUserEval} under both GPT-4-turbo and our Math-Critique for scoring. 

To sum up, our contributions are as follows:

\begin{itemize}[leftmargin=*,itemsep=0pt,parsep=0.2em,topsep=0.2em,partopsep=0.0em]

\item We propose the Self-Critique pipeline, designed to simultaneously enhance mathematical and linguistic abilities. This pipeline operates under the supervision of the Math-Critique model alone, eliminating the need for additional external supervisory models and manual annotations.
We experiment on a ChatGLM3-32B LLM and achieve state-of-the-art across similar-sized LLMs on the \textsc{MathUserEval}, Ape210k~\cite{zhao2020ape210k}, MATH~\cite{hendrycks2020measuring}, and the language subset of AlignBench~\cite{liu2023alignbench}.

\item We introduce the \textsc{MathUserEval} benchmark, designed to evaluate a model's proficiency in open-ended mathematical questions in real-world scenarios.
Compared to existing academic math datasets~\cite{zhao2020ape210k,wang-etal-2017-deep-math23,hendrycks2020measuring,cobbe2021training}, it is more challenging and caters to users' practical needs in application.

\item We analyzed the key factors influencing the enhancement of a model's mathematical abilities using RFT and DPO by Self-Critique pipeline, providing suggestions for future development directions for models to iteratively improve their capabilities independently.
Related techniques from the pipeline have been deployed to GLM-4, helping it to reach performance comparable to GPT-4-0613 across multiple datasets.
\end{itemize}
}

\section{Related Work}
\vpara{LLM for Math Problem-Solving.}
Various approaches have been explored to enhance the mathematical problem-solving abilities of language models. Prompting Methods, initiated by Chain of Thought prompting ~\cite{wei2023chainofthought}, have been refined to guide models through detailed reasoning, with notable contributions from ~\cite{yao2023tree,besta2023graph,yang2023large} enhancing mathematical and reasoning tasks. These methods, however, are limited by the model's inherent capabilities as they do not modify the model itself.
Supervised Fine-tuning and Reinforcement Learning (RL) have also been pivotal. High-quality supervisory data from works like ~\cite{wizardmath,yuan2023scaling-mathrft,abel-math,metamath,yue2023mammoth,zhang2024sciglm} aim to improve model capabilities directly. RL's potential in general domains, demonstrated by ~\cite{openai2023gpt4,touvron2023llama,deepseekai2024deepseek}, has led to mathematical enhancements through OpenAI's Process Reward Model ~\cite{lightman2023lets-verify} and PPO algorithm applications ~\cite{wizardmath,wang2023mathshepherd}, despite the challenges in applying the DPO algorithm ~\cite{rafailov2023direct} for mathematical tasks.
Decoding Strategy and the use of External Tools offer additional avenues for improvement. Self-Consistency ~\cite{wang2023selfconsistency} and MATH-SHEPHERD ~\cite{wang2023mathshepherd} explore sampling strategies to enhance problem-solving while code insertion techniques in works like Mammoth ~\cite{yue2023mammoth} and Tora ~\cite{gou2023tora} bypass computational limitations.

\vpara{Mathematical Evaluation.}
Complex reasoning tasks, such as mathematics, have always been one of the key indicators in assessing the capabilities of language models ~\cite{koncel2016mawps, polu2020generative, hendrycks2021measuring-math, fu2023chain}. Solving a mathematical problem involves semantic understanding, problem decomposition, symbolic reasoning, and numerical computation, making it an unparalleled choice for evaluating the cognitive reasoning ability of LLMs.
The GSM8k ~\cite{cobbe2021training} and MATH ~\cite{hendrycks2021measuring-math} datasets have become the most widely used benchmarks. The GSM8K dataset focuses on multi-step reasoning with basic arithmetic, showcasing the complexity and linguistic diversity inherent in grade school-level mathematics. The MATH benchmark further spans various subjects and difficulty levels, facilitating a detailed examination of a model's problem-solving skills.

Additionally, as one of the most focused-on abilities in language models, there exists a vast array of datasets for mathematical capability across various languages. Some standout sets like AQuA ~\cite{ling2017program} , Mathematics ~\cite{saxton2019analysing} and SAT-Math ~\cite{zhong2023agieval} dive deep into pure math prowess, while NumGLUE ~\cite{mishra2022numglue} and BBH ~\cite{suzgun2022challenging} mix math with other abilities. Moreover, in Chinese, Math23K ~\cite{wang-etal-2017-deep-math23} and CMath ~\cite{wei2023cmath} make elementary and middle school math a playground for problem-solvers, with AgiEval ~\cite{zhong2023agieval} and GaoKaoBench ~\cite{Zhang2023EvaluatingTP-gaokaobench} turn to exam-level challenges. Moreover, Math401 ~\cite{yuan2023large-math401-llm} zeroes in on the ability of number calculation. However, these data are primarily in fixed formats, and current works have found that simple perturbations to these questions can significantly impact performance ~\cite{kumar2021adversarial, zhou2023mathattack}. Therefore, performance on these datasets needs to accurately reflect how models would perform when faced with user math questions.


\begin{figure*}[t]
    \centering
    \includegraphics[width=\linewidth]{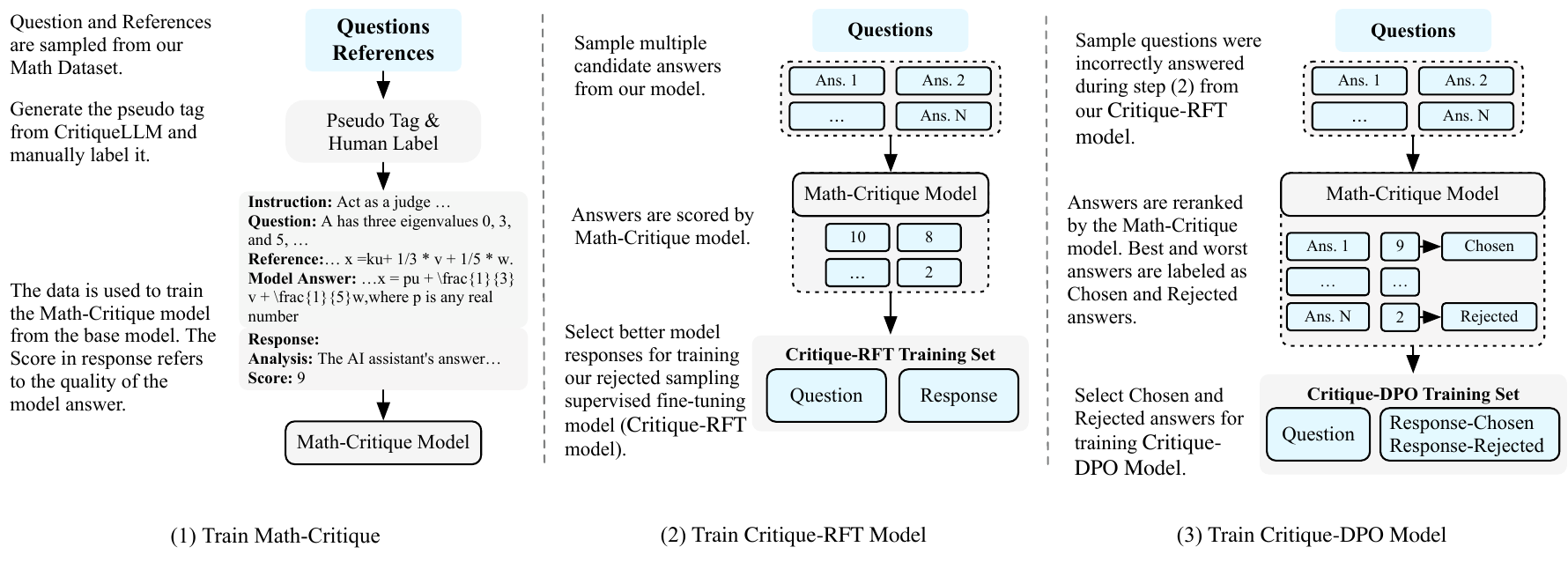}
    \caption{Self-Critique pipeline for ChatGLM-Math. \textmd{Our method comprises three steps: training the Math-Critique model, then utilizing the results from Math-Critique for sampling, followed by two stages of training: Critique RFT  and Critique DPO. Throughout the entire process, only a minimal amount of manual involvement is required during the Math-Critique training phase. Subsequent steps can be fully automated and do not depend on external supervisory models. }} 
\end{figure*}


\section{Math-Critique: A General Critic for Math}

\vpara{Definition.}
The first part of our work involves constructing an accurate and robust evaluation model. We propose Math-Critique, inspired by works that use large models for evaluation purposes~\cite{ke2023critiquellm,zheng2023judging}. This method scores mathematical responses generated by models based on questions and reference answers, including an output of explanatory analysis and a score between 1 and 10. Compared to traditional reward models, this approach leverages the contextual capabilities of language models, enabling more accurate judgments by integrating reference answers. The use of explanatory analysis combined with scoring is also inspired by the concept of thought chains, enhancing scoring accuracy while providing interpretability. 

In the instructions, Math-Critique must classify responses into four categories: entirely incorrect, partially correct methodology yet erroneous outcome, accurate conclusion with partially flawed methodology, and wholly correct. These categories are aligned with scoring ranges of 1–2, 3–5, 6–8, and 9–10, respectively.

Therefore, the Math critique model can be defined as:
$$ \texttt{MathCritique}(\text{Question}, \text{Reference}, \text{Answer}) \rightarrow (\text{Critique}, \text{Score}) $$

Here, the Question and Reference are the original problem and the reference answer, respectively, and the Model Answer is the answer given by the model being evaluated.

We employed two evaluation methods using math-critique: average score evaluation and hard-split evaluation. The average score evaluation calculates the mean of the critique scores assigned to each model answer for a set of questions. On the other hand, the hard-split evaluation categorizes each model answer as either passing or failing based on a predefined correctness threshold. If the critique score surpasses this threshold for each answer, the answer is considered correct; otherwise, it is deemed incorrect. The overall score is then calculated as the proportion of correct answers out of the total number of questions. 



\vpara{Data Collection.}
Our construction method involves the following steps:
\begin{itemize}[leftmargin=*,itemsep=0pt,parsep=0.2em,topsep=0.2em,partopsep=0.0em]
\item  We redesigned the scoring rules and intervals for mathematical responses, enabling the model to grade based on the correctness of the result and the process.
\item  We filtered a dataset from the training data, which includes mathematics questions along with their reference answers and model responses, primarily sourced from exam questions ranging from junior high to university levels. We utilized model sampling answers from multiple sources, including different versions of ChatGLM and other models.
\item  We employed CritiqueLLM~\cite{ke2023critiquellm} and ORM to annotate the dataset, selecting annotations that represented the best and worst scoring extremes from these models, and directly used these \textbf{pseudo tags} for training. This step generated a total of 10k annotated data entries.
\item For results with scores in the middle range, we selected a portion for \textbf{manual annotation} into four categories and then mapped these outcomes to a 10-point scale. We also divided a test set from the training dataset and used the same method for four-category annotation. This step generated 5k annotated data entries for the training set and 800 for the test set.
\end{itemize}

\section{The Self-Critique Pipeline}
\vpara{Overview.}
Based on the construction method of Math-Critique, this section introduces the \textbf{Self-Critique} pipeline. This pipeline is a weakly supervised iterative training method for enhancing mathematical abilities, originating from a single model. Initially, we train a Math-Critique model using the base model and concurrently train a basic Chat Model using the fundamental SFT dataset. Subsequently, we employ the Math-Critique model to supervise the fine-tuning of the Chat Model through rejection sampling. The outcome of this step can serve as a new base model to update both the Math-Critique model and the rejection sampling supervised fine-tuning model. Building upon these steps, our final action involves utilizing the latest Math-Critique model to sample contrast data and then proceeding with DPO training. 


In these steps, the data construction for the Math-critique-base involves a small amount of manual annotation. However, this batch of annotations is a one-time effort, as only this batch of annotated data is needed as a bootstrap for the remaining iterations. After that, inference and automatic model filtering can complete all remaining steps.

Replacing manual annotation with inference can significantly reduce the time required for each iteration from the base model to the final chat model. This avoids a problematic scenario: after weeks of manual annotation, the base model may have undergone further pretraining and become stronger, making it unclear whether the data sampled weeks ago is still accurate or has a significant distribution gap for the current model.

\subsection{Stage 1: Rejective Fine-tuning}
We utilized a rejection sampling method based on Math-Critique.  We re-examined and redesigned the implementation of RFT and found that both the sampling range and the model influence the outcomes during the rejection sampling process. Specifically, we designed the following sampling principles:
\begin{itemize}[leftmargin=*,itemsep=0pt,parsep=0.2em,topsep=0.2em,partopsep=0.0em]
\item  Pre-deduplication: Cluster question embeddings from the training set and evenly sample across categories, ensuring a diverse range of questions without repetition.
\item  Post-sampling deduplication: We conducted a selection process after 5-10 sampling iterations based on the results from Math-Critique. After essential deduplication, we chose the entirely correct response only in cases where there were both correct and incorrect responses to the same question.
\end{itemize}

Following the process outlined above, we have obtained the Critique-RFT dataset:
$$
D_{\text{RFT}} = \left\{ (q_i, a_{ij}) \,|\, \frac{1}{n} \sum_{x} \texttt{MathCritique}(a_{ix}) < 1 \, \text{and } \texttt{MathCritique}(a_{ij}) > \text{correct-bound} \right\}
$$

In this dataset, $q_i$ denotes the ith sampled question, with each question undergoing $n$ samplings. $a_{ij}$ represents the jth response to the ith question. $\texttt{MathCritique}$ refers to Math-Critique score. 'correct bound' denotes the minimum acceptable score for a correct answer, generally set at 0.7.

\subsection{Stage 2: Direct Preference Optimization}

We employed the DPO method to enhance model capabilities further following Critique RFT. The primary advantages of this method are its simplicity in constructing data flows, stability, and speed during training. The DPO method directly compares the correct and incorrect answers to the same question. In our approach, both answers are sampled from the model post-RFT, which we found to be critically important. We also integrated sft loss of DPO positive examples during training as an approximate substitute for a regularization term. 

Our DPO data filtering process is similar to Critique RFT, with the sole difference being the construction method of DPO training pairs. For the selection of DPO pairs, under the premise that there is at least one correct and one incorrect answer, we choose the data pair with the most significant difference in Math-Critique scoring results.

Following the process outlined above, we have obtained the Critique-DPO dataset:


\[
D_{\text{DPO}} = \left\{ (q_i, a_{\text{chosen}}, a_{\text{rejected}}) \middle| 
\begin{array}{l}
\frac{1}{n} \sum_{x} \texttt{MathCritique}(a_{ix}) < 1, \\
\texttt{MathCritique}(a_{\text{chosen}}) > \text{correct-bound}, \\
\texttt{MathCritique}(a_{\text{rejected}}) < \text{rejected-bound}
\end{array} \right\}
\]

In this dataset, each element is a tuple, where $q_i$ is the ith sampled question. For every question $q_i$, sampled $n$ responses, each denoted by $a_{ix}$. The Math-Critique (\texttt{MathCritique}) score is computed for each response $a_{ix}$, and the average of these scores must be less than 1. The chosen answer for each question, $a_{i-chosen}$, is the one that exceeds the 'correct-bound', which is a predetermined threshold indicating a satisfactory level of correctness, often set above a specific value. Conversely, $a_{i-rejected}$ represents the answer that falls below the 'rejected-bound', which is the threshold below which answers are considered incorrect or unsatisfactory.

\subsection{Training}


\subsubsection{Math-Critique Training}
We employ the base model of ChatGLM3-32B~\cite{zeng2022glm,du2022glm} as the initial Math-Critique base model. After each iteration, the model currently refined through SFT (Supervised Finetuning) or Critique RFT will be used as the base. We use a learning rate 3e-6 and a batch size 128 on both 6B and 32B scales.

\subsubsection{Critique-RFT Training}
During the Critique RFT phase, each of our finetuning iterations includes the datasets from previous stages after deduplication, which also encompasses the initial sft dataset. We merge $D_{\text{RFT}}$ and $D_{\text{SFT }}$ as:
$$
D = D_{\text{RFT}} \cup D_{\text{SFT}}
$$
The $D_{\text{SFT }}$ dataset encompasses many routine tasks and can be substituted with an open-source instruction finetuning dataset. To eliminate the potential interference of this dataset on the final results, we compared the impact of including or excluding the sft data in our ablation study. We finetune a base LLM model $\pi_\theta$  by standard max-loglikelihood loss:
\begin{align*}
\mathcal{L}(\pi_\theta) = -\mathbb{E}_{(q_i,a_{ij})\sim \mathcal{D}}\left[  \log \left( \pi_{\theta}(a_{ij} | q_i) \right) \right]
\end{align*}
In this stage, we use a learning rate 2e-5 and finetune for 8000 steps with a batch size of 64.

\begin{figure*}
    \centering
    \includegraphics[width=\linewidth]{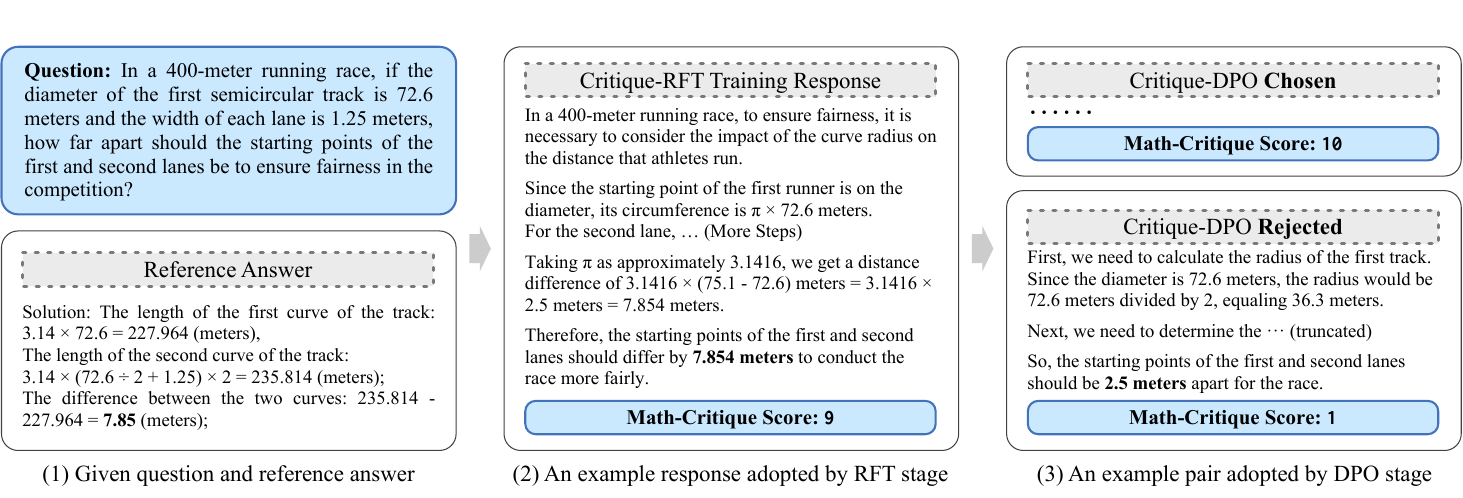}
    \label{fig:main-example}
    \vspace{-5mm}
    \caption{Training datasets examples. \textmd{The training data we generated is divided into two categories, originating from the questions and references within existing datasets. We have constructed separate RFT training data and paired DPO  training data.}
} 
\end{figure*}

\subsubsection{Critique-DPO Training}
During the Critique-DPO phase, it was observed that the direct use of DPO loss led to instability in the training process. A cross-entropy loss for the chosen answer was introduced as a regularization term to the total loss to mitigate this issue. This addition aimed to enhance the stability of the model training. The DPO dataset, previously constructed and denoted as $D_{\text{DPO}}$, was used as the training dataset.
The loss function we used is as follows:
\begin{align*}
& \mathcal{L}_{\rm DPO}(\pi_\theta; \pi_{ref}) = -\mathbb{E}_{(q_i, a_{\rm i-cho}, a_{\rm i-rej})\sim \mathcal{D_{\text{DPO}}}} 
\left[ \log \sigma \left( \beta \log \frac{\pi_\theta(a_{\rm i-cho} | q_i)}{\pi_{ref}(a_{\rm i-cho} | q_i)} - \beta \log \frac{\pi_\theta(a_{\rm i-rej} | q_i)}{\pi_{ref}(a_{\rm i-rej} | q_i)} \right) \right]
\end{align*}
\begin{align*}
\mathcal{L}_{\rm CE}(\pi_\theta; \pi_{ref}) = -\mathbb{E}_{(q_i,a_{\rm i-cho})\sim \mathcal{D}}\left[  \log \left( \pi_{\theta}(a_{\rm i-cho} | q_i) \right) \right]
\end{align*}
\begin{align*}
& \mathcal{L}_{\text{merge}} = \lambda \cdot \mathcal{L}_{\rm DPO} + \mathcal{L}_{\rm CE}
\end{align*}
 In this context, $\lambda$ represents the coefficient of the cross-entropy loss for the chosen answer in the total loss. Commonly, we experiment with values in \{0.5, 1, 1.5\}. Another critical coefficient is $\beta$, which measures the penalty intensity of DPO for incorrect answers. Owing to the addition of a regularization term, the value of this coefficient is higher than that of the standard DPO, with our testing range for this value being \{0.5, 1, 2\}. Besides these, the overall learning rate is set at 1e-6. The experimental section will report the optimal results under these coefficient settings. We train 500 steps with a batch size of 64 in this stage.
 

\begin{table}
\caption{The composition of the \textsc{MathUserEval} data set. \textmd{We divided the test set into three categories: Elementary and Advanced Mathematics. For calculating the total score, we used the macro-average score.}}
\centering
\label{tab:matheval}
\begin{tabular}{@{}llcc@{}}
\toprule
Category                    & Sub-Category   & Size & Source                            \\ \midrule
\multirow{4}{*}{Elementary} & Calculate      & 75   & \multirow{4}{*}{Dialogues}        \\
                            & Algebra        & 113  &                                   \\
                            & Geometry       & 81   &                                   \\
                            & Trigonometry   & 73   &                                   \\ \midrule
\multirow{4}{*}{Advanced}   & Discrete Math  & 45   & \multirow{4}{*}{Dialogues\&Exams} \\
                            & Probability    & 46   &                                   \\
                            & Linear Algebra & 58   &                                   \\
                            & Calculus       & 54   &                                   \\ \bottomrule
\end{tabular}
\vspace{2mm}

\end{table}

\section{\textsc{MathUserEval}: Benchmarking LLMs' Mathematical Reasoning in Application}
\textsc{MathUserEval} is a test set designed for real-use scenarios, targeting issues of concern to users and more challenging mathematical problems. Some of our data originates from university examination questions, while another is from simulated dialogues. For the latter, we assigned a series of annotators who posed math-related questions to our system based on their daily experiences and observations using large models.

Based on the distribution of the collected data, we divided the test set into two main categories, Elementary and Advanced, and eight sub-categories. Given that Calculate  Applications are less challenging and closely aligned with the scope of previous public datasets, we selected fewer questions from this category.  The quantity of questions in each of these categories is as shown in Table~\ref{tab:matheval}. All questions are posed in an open-ended format. Possible answers include a single number, multiple numbers, or mathematical expressions.

We offer two evaluation methods: GPT-4-1106——Preview~\cite{openai2023gpt4,liu2023alignbench,zheng2023judging} evaluation and Math-Critique evaluation. The former adopts the evaluation method of alignbench~\cite{liu2023alignbench}, to provide a more accurate, fair, and accessible evaluation approach; the latter employs the same usage as the Math-Critique introduced above. Similarly, we will also report two types of scores: avg-score and hard-split.
\section{Experiment}
\subsection{Data Collection}

The primary sources of our data collection can be categorized as follows: training sets from public datasets and publicly available middle school and university examination questions. We have selected all prompts from  GSM8k ~\cite{cobbe2021training} and MATH ~\cite{hendrycks2021measuring-math} training set as the question set for the English data, and used the responses from the original dataset as the standard answers. Regarding publicly available middle school and university exam questions, we used the answer formats provided with the exam papers as the common answers without further processing.

\begin{table*}[t]
\caption{Main Result. \textmd{All results reported are the highest achieved in zero-shot or few-shot settings and are based on greedy decoding. The best models are marked in \textbf{bold} and the {\ul underline} signifies the second best model.}}
\vspace{-3mm}
\label{table:main-result}
\centering

\renewcommand\tabcolsep{0.7pt}
\renewcommand\arraystretch{1.1}
\resizebox{\textwidth}{!}{
\begin{threeparttable}
\begin{tabular}{@{}lccccccccccc@{}}
\toprule
\multirow{3}{*}{Models}                           & \multirow{3}{*}{\#params} & \multicolumn{5}{c}{Chinese}                                                                       & \multicolumn{3}{c}{English}                                                                                             & \multicolumn{2}{c}{General}       \\ \cmidrule(lr){3-7} \cmidrule(lr){8-10} \cmidrule(l){11-12} 
                                                  &                           & \multicolumn{3}{c}{MathUserEval}              & \multirow{2}{*}{Ape210k} & \multirow{2}{*}{Cmath} & \multirow{2}{*}{GSM8k} & \multirow{2}{*}{MATH} & \multirow{2}{*}{\begin{tabular}[c]{@{}c@{}}Hunga\\ -rian\end{tabular}} & AlignBench & \multirow{2}{*}{MT-Bench}    \\ \cmidrule(lr){3-5} \cmidrule(l){11-11} 
                                                  &                           & Overall       & Elementary    & Advanced      &                          &                        &                        &                       &                                                                        & Language      \\ \midrule
\footnotesize GPT-4-1106-Preview~\cite{openai2023gpt4}          & N/A                       & \textbf{5.73} & \textbf{5.07} & \textbf{6.81} & {\ul 84.2}               & \textbf{89.3}          & \textbf{93.6}          & \textbf{53.6}         & \textbf{92}                                                            & {\ul 8.29}  & \textbf{9.32}  \\
\footnotesize GPT-4-0613~\cite{openai2023gpt4}                  & N/A                       & 4.14          & 3.34          & 5.33          & 83.6                     & 86.5                   & 91.4                   & 45.8                  & 68                                                                     & 7.59  & {\ul9.18}        \\
\footnotesize GPT-3.5-Turbo-0613~\cite{openai2023gpt4}          & N/A                       & 3.42          & 3.04          & 4.07          & 70.4                     & 76.8                   & 78.2                   & 28.0                  & 41                                                                     & 6.82 &8.36         \\
\footnotesize Claude-2~\cite{Claude}                            & N/A                       & 3.29          & 2.63          & 4.35          & 72.8                     & 80.5                   & 88.0                   & -                     & 55                                                                     & 6.78  & 8.06        \\
\footnotesize GLM-4                                             & N/A                       & {\ul 5.11}    & {\ul 4.86}    & {\ul 5.43}    & \textbf{93.5}            & {\ul 89.0}             & {\ul 91.8}             & {\ul 49.0}            & {\ul 75}                                                               & \textbf{8.38} & 8.62\\ \midrule
\footnotesize Skywork-13B-Math~\cite{yang2023skymath}           & 13B                       & 2.66          & 2.75          & 2.54          & 74.4                     & 77.3                   & 72.3                   & 17.0                  & 39                                                                     & 5.58        & 4.12  \\
\footnotesize InternLM2-Chat~\cite{2023internlm}                & 20B                       & 3.25          & 3.00          & 3.68          & 72.0                     & 80.7                   & 79.6                   & 34.8                  & 48                                                                     & {\ul 7.68} & {\ul8.21}   \\
\footnotesize Math-InternLM2~\cite{2023internlm}                & 20B                       & 3.17          & 3.08          & 3.37          & 75.2                     & 78.5                   & {\ul 82.6}             & 37.7                  & {\ul 66}                                                               & 6.53        & 6.09  \\
\footnotesize Yi-Chat~\cite{Yi}                                 & 34B                       & 2.64          & 2.49          & 2.87          & 65.1                     & 77.7                   & 76.0                   & 15.9                  & 39                                                                     & 6.18      & 6.54    \\
\footnotesize DeepSeek-Chat~\cite{deepseekai2024deepseek}      & 67B                       & 3.24          & 2.76          & 3.84          & 76.7                     & 80.3                   & \textbf{84.1}          & 32.6                  & 58                                                                     & 7.11       & \textbf{8.35}   \\
\footnotesize MetaMath (EN)~\cite{metamath}                     & 70B                       & -             & -             & -             & -                        & -                      & 82.3                   & 26.0                  & 35                                                                     & -            & 4.28 \\
\footnotesize Qwen-Chat~\cite{bai2023qwen}                      & 72B                       & 3.87          & {\ul 3.99}    & 3.67          & 77.1                     & \textbf{88.1}          & 76.4                   & 31.8                  & 52                                                                     & 7.29       & 6.43   \\
\rowcolor[HTML]{DAE8FC} \footnotesize{ChatGLM3-32B-SFT-2312$^*$}      & 32B                       & 3.25          & 3.03          & 3.60          & 78.0                     & 79.8                   & 75.8                   & 29.0                  & 39                                                                     & 7.37       & 8.05   \\
\rowcolor[HTML]{DAE8FC} \footnotesize{\qquad\qquad\qquad   + RFT}      & 32B                       & {\ul 4.01}    & 3.86          & {\ul 4.26}    & {\ul 87.0}               & 85.3                   & 82.4                   & {\ul 39.5}            & 58                                                                     & 7.42       & 8.03   \\
\rowcolor[HTML]{DAE8FC} \footnotesize{\qquad\qquad\qquad   + RFT, DPO} & 32B                       & \textbf{4.23} & \textbf{4.01} & \textbf{4.59} & \textbf{89.4}            & {\ul 85.6}             & {\ul 82.6}             & \textbf{40.6}         & \textbf{73}                                                            & \textbf{7.80} & 8.08 \\ \bottomrule
\end{tabular}
\begin{tablenotes}
\item[*] ChatGLM3-32B-SFT-2312 is a newer version of the ChatGLM series and not identical to the model discussed in~\cite{hou2024chatglmrlhf}, despite sharing the same model size.
\end{tablenotes}
\end{threeparttable}
}
\vspace{-5mm}

\end{table*}

\hide{

\begin{table*}[t]
\footnotesize
\centering
\resizebox{\textwidth}{!}{
\begin{tabular}{@{}lccccccccc@{}}
\toprule
\multicolumn{1}{c}{}                         &                            & \multicolumn{5}{c}{Chinese Datasets}                                                                & \multicolumn{3}{c}{English Datasets}                                           \\ \cmidrule(l){3-7} \cmidrule(l){8-10}
\multicolumn{1}{c}{}                         &                            & \multicolumn{3}{c}{MathUserEval}              &                           &                         &                         &                        &                             \\ \cmidrule(lr){3-5}
\multicolumn{1}{c}{\multirow{-3}{*}{Models}} & \multirow{-3}{*}{\#params} & Overall       & Elementary    & Advanced      & \multirow{-2}{*}{Ape210k} & \multirow{-2}{*}{Cmath} & \multirow{-2}{*}{GSM8k} & \multirow{-2}{*}{MATH} & \multirow{-2}{*}{Hungarian} \\ \midrule
\multicolumn{10}{c}{Closed LLM APIs}                                                                                                                                                                                                                             \\ \midrule
gpt-4-1106-preview                           & N/A                        & \textbf{5.73} & \textbf{5.07} & \textbf{6.81} & 84.2                      & 89.3                    & \textbf{93.6}           & \textbf{53.6}          & \textbf{92}                 \\
gpt-4-0613                                   & N/A                        & 4.14          & 3.34          & 5.33          & 83.6                      & 86.5                    & 91.4                    & 45.8                   & 68                          \\
gpt-3.5-turbo-0613                           & N/A                        & 3.42          & 3.04          & 4.07          & 70.4                      & 76.8                    & 78.2                    & 28.0                   & 41                          \\
Claude-2                                     & N/A                        & 3.29          & 2.63          & 4.35          & 72.8                      & 80.5                    & 88                      & -                      & 55                          \\
ERNIE 4.0                                    & N/A                        & 4.83          & 4.50          & 5.31          & {\ul 90.2}                & \textbf{94.0}           & 86.2                    & {\ul 51.6}             & -                           \\
Qwen-Max                                     & N/A                        & 4.38          & 4.07          & 4.84          & 88.7                      & 88.6                    & 78.6                    & -                      & -                           \\
\rowcolor[HTML]{DAE8FC} 
GLM4                                         & N/A                        & {\ul 5.11}    & {\ul 4.86}    & {\ul 5.43}    & \textbf{93.5}             & {\ul 89.0}              & {\ul 91.8}              & 49.0                   & {\ul 71}                    \\ \midrule
\multicolumn{10}{c}{Medium-sized LLMs (11B+)}                                                                                                                                                                                                                    \\ \midrule
Skywork-13B-Math                             & 13B                        & 2.66          & 2.75          & 2.54          & 74.4                      & 77.3                    & 72.3                    & 17.0                   & 39                          \\
InternLM2-Chat                               & 20B                        & 3.25          & 3.00          & 3.68          & 72                        & 80.7                    & 79.6                    & 34.8                   & 48                          \\
Math-InternLM2                               & 20B                        & 3.17          & 3.08          & 3.37          & 75.2                      & 78.5                    & {\ul 82.6}              & 37.7                   & \textbf{66}                 \\
Yi-Chat                                      & 34B                        & 2.64          & 2.49          & 2.87          & 65.1                      & 77.7                    & 76                      & 15.9                   & 39                          \\
DeepSeek-Chat                               & 67B                        & 3.24          & 2.76          & 3.84          & 76.7                      & 80.3                    & \textbf{84.1}           & 32.6                   & {\ul 58}                    \\
MetaMath(EN)                                 & 70B                        & -             & -             & -             & -                         & -                       & 82.3                    & 26.0                   & 35                          \\
Qwen-Chat                                    & 72B                        & 3.87          & {\ul 3.99}    & 3.67          & 77.1                      & \textbf{88.1}           & 76.4                    & 31.8                   & 52                          \\
\rowcolor[HTML]{DAE8FC} 
ChatGLM3-32B-SFT-2312                                  & 32B                        & 3.39          & 3.35          & 3.44          & 85.9                      & 83.3                    & 78.2                    & 34.5                   & 56                          \\
\rowcolor[HTML]{DAE8FC} 
\qquad\qquad   + RFT                        & 32B                        & {\ul 4.01}    & 3.86          & {\ul 4.26}    & {\ul 87.0}                & 85.3                    & 82.4                    & {\ul 39.5}             & {\ul 58}                    \\
\rowcolor[HTML]{DAE8FC} 
\qquad\qquad   + RFT, DPO                   & 32B                        & \textbf{4.23} & \textbf{4.01} & \textbf{4.59} & \textbf{89.4}             & {\ul 85.6}              & {\ul 82.6}              & \textbf{40.6}          & \textbf{66}                 \\ \bottomrule
\end{tabular}
}
\vspace{1mm}
\caption{Main Result. \textmd{All results reported are the highest achieved in zero-shot or few-shot settings and are based on greedy decoding. In the classification of Close-source and Open-source models, the best models are marked in \textbf{bold}, respectively. Within the Close-source category, the {\ul underline} signifies the best model excluding gpt-4-1106-preview.}}
\vspace{-5mm}
\label{table:main-result}
\end{table*}

\begin{table*}[t]
\resizebox{\textwidth}{!}{
\renewcommand\tabcolsep{4pt}
\begin{tabular}{@{}lccccccccc@{}}
\toprule
\multicolumn{1}{c}{\multirow{3}{*}{Models}} & \multirow{3}{*}{\#params} & \multicolumn{5}{c}{Chinese Test Set}                                                              & \multicolumn{3}{c}{English Test Set}                                    \\ \cmidrule(l){3-10} 
\multicolumn{1}{c}{}                        &                           & \multicolumn{3}{c}{MathUserEval}              & \multirow{2}{*}{Ape210k} & \multirow{2}{*}{Cmath} & \multirow{2}{*}{GSM8k} & \multirow{2}{*}{MATH} & \multirow{2}{*}{Hung.} \\
\multicolumn{1}{c}{}                        &                           & Overall       & Elementary    & Advanced      &                          &                        &                        &                       &                        \\ \midrule
\multicolumn{9}{c}{Close-source models}                                                                                                                                                                                      &                        \\
gpt-4-1106-preview~\cite{openai2023gpt4}                          & N/A                       & \textbf{5.73} & \textbf{5.07} & \textbf{6.81} & 84.2                     & 89.3                   & \textbf{93.6}          & \textbf{53.6}         & \textbf{92}            \\
gpt-4-0613~\cite{openai2023gpt4}                                  & N/A                       & 4.14          & 3.34          & 5.33          & 83.6                     & 86.5                   & 91.4                   & 45.8                  & 68                     \\
gpt-3.5-turbo-0613~\cite{openai2023gpt4}                          & N/A                       & 3.42          & 3.04          & 4.07          & 70.4                     & 76.8                   & 78.2                   & 28.0                  & 41                     \\
Claude-2~\cite{Claude}                                    & N/A                       & 3.29          & 2.63          & 4.35          & 72.8                     & 80.5                   & 88.0                   & -                     & 55                     \\
ERNIE4.0~\cite                                    & N/A                       & 4.83          & 4.50          & 5.31          & 90.2                     & {\ul \textbf{94.0}}    & 86.2                   & {\ul \textbf{51.6}}   & -                      \\
Qwen-Max~\cite{bai2023qwen}                                    & N/A                       & 4.38          & 4.07          & 4.84          & 88.7                     & 88.6                   & 78.6                   & -                     & -                      \\
\textbf{GLM4(Ours)}                         & N/A                       & {\ul 5.11}    & {\ul 4.86}    & {\ul 5.43}    & {\ul \textbf{93.5}}      & 89.0                   & {\ul 91.8}    & 49.0                  & {\ul 75}      \\ \midrule
\multicolumn{9}{c}{Open-source models (11B+)}                                                                                                                                                                                &                        \\
Skywork-13B-Math~\cite{yang2023skymath}                            & 13B                       & 2.66          & 2.75          & 2.54          & 74.4                     & 77.3                   & 72.3                   & 17.0                  & 39                     \\
InternLM2-Chat~\cite{2023internlm}                              & 20B                       & 3.25          & 3.00          & 3.68          & 72.0                     & 80.7                   & 79.6                   & 34.8                  & 48                     \\
Math-InternLM2~\cite{2023internlm}                              & 20B                       & 3.17          & 3.08          & 3.37          & 75.2                     & 78.5                   & 82.6                   & 37.7                  & 66            \\
Yi-Chat~\cite{Yi}                                     & 34B                       & 2.64          & 2.49          & 2.87          & 65.1                     & 77.7                   & 76.0                   & 15.9                  & 39                     \\
DeepSeek-Chat~\cite{deepseekai2024deepseek}                              & 67B                       & 3.24          & 2.76          & 3.84          & 76.7                     & 80.3                   & \textbf{84.1}          & 32.6                  & 58                     \\
MetaMath(EN)~\cite{metamath}                                & 70B                       & -             & -             & -             & -                        & -                      & 82.3                   & 26.0                  & 35                     \\
Qwen-Chat~\cite{bai2023qwen}                                   & 72B                       & 3.87          & 3.99          & 3.67          & 77.1                     & \textbf{88.1}          & 76.4                   & 31.8                  & 52                     \\
\textbf{GLM-32b-SFT w/o RFT\&DPO}           & 32B                       & 3.39          & 3.35          & 3.44          & 85.9                     & 83.3                   & 78.2                   & 34.5                  & 56                     \\
\textbf{GLM-Math-32b w/o DPO}               & 32B                       & 4.01          & 3.86          & 4.26          & 87.0                     & 85.3                   & 82.4                   & 39.5                  & 58                     \\
\textbf{GLM-Math-32b w/ DPO}                & 32B                       & \textbf{4.23} & \textbf{4.01} & \textbf{4.59} & \textbf{89.4}            & 85.6                   & 82.6                   & \textbf{40.6}         & \textbf{73}            \\ \bottomrule
\end{tabular}
}
\vspace{1mm}
\caption{Main Result. \textmd{All results reported are the highest achieved in zero-shot or few-shot settings and are based on greedy decoding. In the classification of Close-source and Open-source models, the best models are marked in \textbf{bold}, respectively. Within the Close-source category, the {\ul underline} signifies the best model excluding gpt-4-1106-preview.}}
\label{table:main-result}
\end{table*}
}

\hide{
\begin{table*}[t]
\resizebox{\textwidth}{!}{
\renewcommand\tabcolsep{4pt}
\begin{tabular}{@{}lccccccccc@{}}
\toprule
\multicolumn{1}{c}{\multirow{3}{*}{Models}} & \multirow{3}{*}{\#params} & \multicolumn{5}{c}{Chinese Test Set}                                                              & \multicolumn{3}{c}{English Test Set}                                    \\ \cmidrule(l){3-10} 
\multicolumn{1}{c}{}                        &                           & \multicolumn{3}{c}{MathUserEval}              & \multirow{2}{*}{Ape210k} & \multirow{2}{*}{Cmath} & \multirow{2}{*}{GSM8k} & \multirow{2}{*}{MATH} & \multirow{2}{*}{Hung.} \\
\multicolumn{1}{c}{}                        &                           & Overall       & Elementary    & Advanced      &                          &                        &                        &                       &                        \\ \midrule
\multicolumn{9}{c}{Close-source models}                                                                                                                                                                                      &                        \\
gpt-4-1106-preview~\cite{openai2023gpt4}                          & N/A                       & \textbf{5.73} & \textbf{5.07} & \textbf{6.81} & 84.2                     & 89.3                   & \textbf{93.6}          & \textbf{53.6}         & \textbf{92}            \\
gpt-4-0613~\cite{openai2023gpt4}                                  & N/A                       & 4.14          & 3.34          & 5.33          & 83.6                     & 86.5                   & 91.4                   & 45.8                  & 68                     \\
gpt-3.5-turbo-0613~\cite{openai2023gpt4}                          & N/A                       & 3.42          & 3.04          & 4.07          & 70.4                     & 76.8                   & 78.2                   & 28.0                  & 41                     \\
Claude-2~\cite{Claude}                                    & N/A                       & 3.29          & 2.63          & 4.35          & 72.8                     & 80.5                   & 88.0                   & -                     & 55                     \\
ERNIE4.0~\cite                                    & N/A                       & 4.83          & 4.50          & 5.31          & 90.2                     & {\ul \textbf{94.0}}    & 86.2                   & {\ul \textbf{51.6}}   & -                      \\
Qwen-Max~\cite{bai2023qwen}                                    & N/A                       & 4.38          & 4.07          & 4.84          & 88.7                     & 88.6                   & 78.6                   & -                     & -                      \\
\textbf{GLM4(Ours)}                         & N/A                       & {\ul 5.11}    & {\ul 4.86}    & {\ul 5.43}    & {\ul \textbf{93.5}}      & 89.0                   & {\ul 91.8}    & 49.0                  & {\ul 75}      \\ \midrule
\multicolumn{9}{c}{Open-source models (11B+)}                                                                                                                                                                                &                        \\
Skywork-13B-Math~\cite{yang2023skymath}                            & 13B                       & 2.66          & 2.75          & 2.54          & 74.4                     & 77.3                   & 72.3                   & 17.0                  & 39                     \\
InternLM2-Chat~\cite{2023internlm}                              & 20B                       & 3.25          & 3.00          & 3.68          & 72.0                     & 80.7                   & 79.6                   & 34.8                  & 48                     \\
Math-InternLM2~\cite{2023internlm}                              & 20B                       & 3.17          & 3.08          & 3.37          & 75.2                     & 78.5                   & 82.6                   & 37.7                  & 66            \\
Yi-Chat~\cite{Yi}                                     & 34B                       & 2.64          & 2.49          & 2.87          & 65.1                     & 77.7                   & 76.0                   & 15.9                  & 39                     \\
DeepSeek-Chat~\cite{deepseekai2024deepseek}                              & 67B                       & 3.24          & 2.76          & 3.84          & 76.7                     & 80.3                   & \textbf{84.1}          & 32.6                  & 58                     \\
MetaMath(EN)~\cite{metamath}                                & 70B                       & -             & -             & -             & -                        & -                      & 82.3                   & 26.0                  & 35                     \\
Qwen-Chat~\cite{bai2023qwen}                                   & 72B                       & 3.87          & 3.99          & 3.67          & 77.1                     & \textbf{88.1}          & 76.4                   & 31.8                  & 52                     \\
\textbf{GLM-32b-SFT w/o RFT\&DPO}           & 32B                       & 3.39          & 3.35          & 3.44          & 85.9                     & 83.3                   & 78.2                   & 34.5                  & 56                     \\
\textbf{GLM-Math-32b w/o DPO}               & 32B                       & 4.01          & 3.86          & 4.26          & 87.0                     & 85.3                   & 82.4                   & 39.5                  & 58                     \\
\textbf{GLM-Math-32b w/ DPO}                & 32B                       & \textbf{4.23} & \textbf{4.01} & \textbf{4.59} & \textbf{89.4}            & 85.6                   & 82.6                   & \textbf{40.6}         & \textbf{73}            \\ \bottomrule
\end{tabular}
}
\vspace{1mm}
\caption{Main Result. \textmd{All results reported are the highest achieved in zero-shot or few-shot settings and are based on greedy decoding. In the classification of Close-source and Open-source models, the best models are marked in \textbf{bold}, respectively. Within the Close-source category, the {\ul underline} signifies the best model excluding gpt-4-1106-preview.}}
\label{table:main-result}
\end{table*}
}
\subsection{Evaluation Setting}
\subsubsection{Datasets} In our research, we primarily tested the \textsc{MathUserEval} dataset, which originates from simulated dialogue records and actual exam papers. Compared to academic datasets, this dataset features a more diverse array of question styles and more closely aligns with real-world usage scenarios. Additionally, we tested the following academic datasets: 1. English academic datasets: GSM8k~\cite{cobbe2021training} and MATH~\cite{hendrycks2021measuring-math}. These two datasets contain English mathematics problems at the middle and high school and competition levels. 2. Chinese academic datasets:  ape210k~\cite{zhao2020ape210k} and cmath~\cite{wei2023cmath}. The questions in these datasets also originate from middle and high school levels. We also employed the Hungarian National Exam~\cite{testing_language_models_on_a_held_out_high_school_national_finals_exam} as an Out-Of-Distribution test set. It should be noted that, across all test sets, we only used the training sets of GSM8k and MATH as seed data for data generation.

To evaluate general linguistic capabilities, we selected the Chinese language component of AlignBench~\cite{liu2023alignbench} and full MT-Bench~\cite{zheng2023judging} for testing.

\subsubsection{Baselines} 
Since most of our work is conducted in Chinese, we selected three categories of baselines: open-source mathematics-specific models, open-source Chinese models, and leading proprietary models. For the open-source mathematics models, we chose SkyMath~\cite{yang2023skymath}, MetaMath~\cite{metamath}, and Internlm2-Math~\cite{2023internlm} as our baselines. To effectively compare with the best Chinese models, we selected Qwen-Chat~\cite{bai2023qwen}, Yi-Chat~\cite{Yi}, DeepSeek-Chat~\cite{deepseekai2024deepseek}, and InternLM2~\cite{2023internlm}. Additionally, we also report the results for GPT-4-1106-Preview(known as GPT4-Turbo), GPT-4-0613,GPT-3.5-Turbo~\cite{openai2023gpt4}, and Claude-2~\cite{Claude}.

\subsubsection{Metrics} For all datasets, we utilized the results of greedy inference performed once. Regarding academic datasets, we report the self-reported results of corresponding models and the highest zero-shot/few-shot results from the OpenCompass and \textsc{MathUserEval} websites. For the math subset of AlignBench~\cite{liu2023alignbench} and our proposed \textsc{MathUserEval} test set, we report the scoring results from GPT-4-Turbo and the scores generated by Math-Critique. More information about evaluation settings can be found in Appendix~\ref{appendix:evaluation}.

\subsection{Main Results}


Table~\ref{table:main-result}  displays our main results. In models with more than 10 billion parameters, our model achieved a score of 4.23 on \textsc{MathUserEval}, 89.4 on ape210k~\cite{zhao2020ape210k}, and 40.6 on MATH~\cite{hendrycks2020measuring}, surpassing all models with published parameters and achieved near-top performances on Cmath and GSM8k. Our model also scored 73 in the Hungary Test~\cite{testing_language_models_on_a_held_out_high_school_national_finals_exam}, the highest score among all known parameter models.

We used the ChatGLM3-32B-SFT-2312 version as our baseline. Our RFT phase significantly improved across all math datasets. In contrast, the DPO  phase's improvement focused on open-ended math problems, including \textsc{MathUserEval}, the Hungarian Exam, and the general-purpose AlignBench. Even though our improvement on MT-bench~\cite{zheng2023judging} is not significant, given that over 90\% of our training data is in Chinese, we believe that maintaining parity essentially demonstrates that our method has preserved the original English general or multi-turn capabilities.

Compared to proprietary models, especially the GPT series by OpenAI, GLM-4 demonstrates competitive or superior performance in specific areas. GPT-4-1106-Preview, for example, shows the best performance in most tasks, including the highest scores in both Chinese and English benchmarks, highlighting its effectiveness in various mathematical problem-solving contexts. However, GLM-4 surpasses it in the Ape210k and AlignBench benchmarks, suggesting particular strengths in mathematical reasoning and cross-linguistic generalization.



\subsection{Ablation Study}
\begin{table}[t]
\centering
\caption{Ablation Study for 32B model. \textmd{All results are fine-tuned from our 32B base model. We selected Metamath training set as baselines that we consider comparatively strong. \textsc{MathUserEval} is scored with Math-Critique model.}}
\label{tab:32b-ablation}
\begin{threeparttable}
\begin{tabular}{@{}lcccc@{}}

\toprule
\multirow{2}{*}{Method}      & \multicolumn{2}{c}{Chinese} & \multicolumn{2}{c}{English} \\ \cmidrule(l){2-5} 
                             & \textsc{MathUserEval}      & Ape210k$^*$      & GSM8k           & MATH$^*$           \\ \midrule
Metamath~\cite{metamath}     & 2.80              & 75.8             & 77.9            & 35.6               \\ \midrule
ChatGLM3-32B-SFT-2312 + RFT          & 3.74              & 87.0             & 82.4            & 39.5               \\
- Real scenarios \& Academic & 3.29              & 85.9             & 74.8            & 27.6               \\
- Real scenarios             & 3.29              & 74.6             & 77.4            & 36.0               \\
- Academic                   & 3.72              & 75.8             & 81.0            & 36.2               \\ \midrule
ChatGLM3-32B-SFT-2312 + RFT \& DPO         & 4.37	& 89.4	& 82.6	& 41.0               \\
- Real Scenarios \& Academic  & 4.14 	& 87.8&  	81.5 & 	37.8  \\ \bottomrule
\end{tabular}
\begin{tablenotes}
\item[*] Ablated experiments are conducted on 500-sample test subsets.
\end{tablenotes}
\end{threeparttable}
\vspace{1mm}

\end{table}

\vpara{Impact of data composition.}
Table~\ref{tab:32b-ablation} presents the results of ablation experiments. We selected Metamath~\cite{metamath} training set as baselines that we consider comparatively strong. After applying Critique-RFT, we found that using only academic datasets to construct RFT data results in inferior performance in real-life scenario-based \textsc{MathUserEval} and academic test sets compared to the outcomes after integrating real-life scenario data. Furthermore, introducing English data significantly improves performance on English datasets without substantially affecting Chinese capabilities.

The ablation experiments conducted during the Critique-DPO phase indicate that, compared to using general DPO data, the addition of math-specific DPO data significantly enhances mathematical capabilities. We no longer conduct separate tests on the impact of Real scenarios and Academic data on capabilities. This is because, in previous stages, questions that the model could solve correctly were removed, making it impossible to ensure that both datasets still have sufficient size for a complete training session.


\vpara{Relationship between Different Boosting Methods and Problem Difficulty.}

\begin{figure}[t]
    \centering
    \includegraphics[width=250pt]{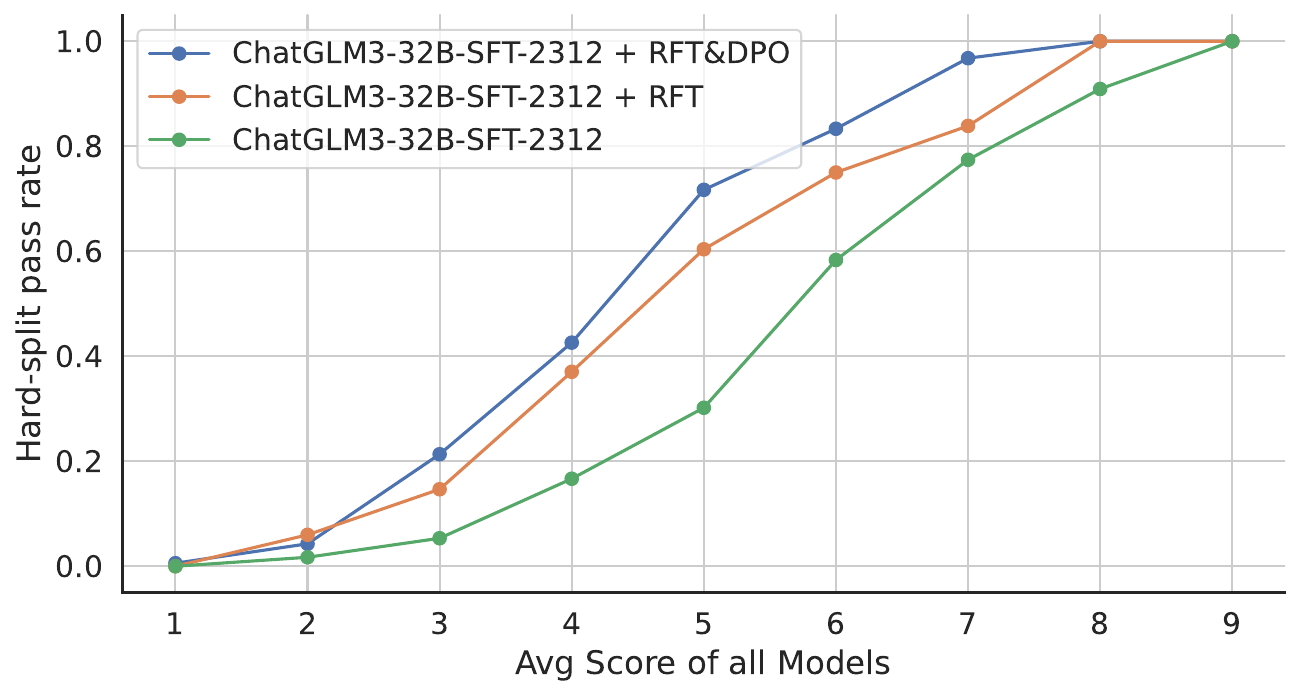}
    \caption{The Relationship between Different Boosting Methods and Problem Difficulty. \textmd{The horizontal axis displays the average score of \textsc{MathUserEval} across 24 models (scored by GPT-4-1106-Preview), which we regard as a representation of problem difficulty. The vertical axis represents the hard-split scores of the models on these questions.}}
    \label{fig:avg-hard split}
    \vspace{-3mm}
\end{figure}
Figure~\ref{fig:avg-hard split} displays the relationship between the average accuracy of each question in \textsc{MathUserEval} across all 24 models tested (including some intermediate models) and the hard-split scores of the four GLM series models. The average accuracy across all tested models is considered a reflection of the difficulty level of the question. It can be observed that the RFT step improves performance across almost all difficulty levels, but the most significant improvements come from questions with an average score between 4 and 6. The DPO step mainly enhances performance on questions with an average score between 5 and 7. This suggests that our two-step approach to enhancing mathematical capabilities can be seen as aligning the model more closely with real-world conditions, with the most noticeable improvements on medium-difficulty questions. 



\vpara{Impact on general capabilities.}
Considering that our goal is not to develop a specialized mathematical model for leaderboard climbing but rather a general model with strong mathematical capabilities, we tested the results using Alignbench~\cite{liu2023alignbench}, a Chinese general open question-answering dataset. The results in Table~\ref{table:main-result} demonstrate that our model exceeds the training outcomes of similar baseline models that do not incorporate specialized mathematical data regarding Chinese language capabilities. Additionally, it performs exceptionally well compared to other open-source Chinese mathematical/general models.

In terms of English general capabilities, we tested using MT-Bench~\cite{zheng2023judging} as the test set. Given that over 90\% of our training data consists of Chinese, the fact that the results on MT-Bench~\cite{zheng2023judging} remained largely unchanged during our training process indicates that the English language capabilities were not significantly affected.


\begin{table}[t]
\centering
\caption{Evaluation for Math-Critique Model.\textmd{~We report "Acc" as the accuracy of the model in determining whether an answer is correct, as well as the Pearson, Spearman, and Kendall correlation coefficients for Math-Critique in comparison with human annotations in a four-category classification.}}
\label{tab:math-critique-eva}
\begin{tabular}{@{}lcccc@{}}
\toprule
Model             & Acc. & Pearson & Spearman & Kendall \\ \midrule
GPT-3.5-Turbo     & 62.1                    & 31.8    & 33.5     & 30.1    \\
GPT-4-0613             & 90.2                    & 80.5    & 78.1     & 71.0    \\
Math-Critique-32B & 90.5                    & 80.4    & 77.1     & 70.2    \\ \bottomrule
\end{tabular}
\vspace{2mm}

\end{table}
\vpara{Effectiveness of Math-Critique.} During the process of manual annotation, we collected a test set of 800 questions, all of which were manually marked for the correctness of their answers and procedures, thus forming a four-category test; the output results of Math-Critique were mapped to these four categories according to the requirements of the instructions.

We validated the effectiveness of Math-Critique itself through empirical experiments. We set up two evaluation methods: the accuracy of directly scoring to judge correct/incorrect results and the accuracy of judging our defined four categories. We extracted test sets from Chinese junior and senior high school exam questions and \textsc{MathUserEval}, annotated correct judgment by experts. 

The results shown in Table~\ref{tab:math-critique-eva} indicate that our Math-Critique-32B model significantly surpasses GPT-3.5-Turbo in both judgment accuracy and correlation coefficients compared to human annotations and is essentially on par with GPT-4-0613.

\vpara{Out-Of-Distribution Test.}
Following the approach of Grok-1, to test the performance on Out-Of-Distribution datasets, we selected the Hungarian national final exam~\cite{testing_language_models_on_a_held_out_high_school_national_finals_exam}. This is a test set of 33 exam questions without a training set, the advantage being that it allows for evaluating a model's mathematical capabilities in an utterly OOD environment. As shown in Figure~\ref{fig:hungarg test}, using human expert evaluation, we found that at a model scale of 32B, our RFT model scored 57, while the DPO model scored 73. However, it is essential to note that since our model's primary language is Chinese, if the model answers correctly in Chinese, we would score it usually. We plan to address this issue in future models.

\section{Limitation and Future Work}

We observed the following issues in our mathematical models, and we leave it for our future work:

\vpara{Graphic thinking and drawing abilities.} Due to the limitations of being a purely linguistic model, our model has deficiencies in handling questions requiring drawing. For example, in a question from the Hungary Test, which required connecting six numbers as divisors of each other, our model correctly listed the different numbers' connecting topology but could not draw it accurately. Also, as a language model, it struggles to respond correctly to questions requiring an understanding of images. A potential solution could be integrating multimodal input and output components, an area we plan to explore further.

\vpara{Precision calculation capability.} We observed that in incorrectly answered questions, if the problem required multiplication, division, or exponentiation of three or more decimal places, our model might compute with a deviation of up to 5\%. This phenomenon aligns with observations from GPT-4 models without an integrated code interpreter. This issue might be a fundamental problem to pure language models and could be mitigated but not resolved with increasing model size. Using external tools for computation or directly employing code with a code interpreter could solve this problem. However, our discussion in this paper focuses on enhancing the mathematical capabilities of pure language models, and we will endeavor to address these issues in future work.

\section{Conclusion}
In this paper, we introduce the Math-Critique method for evaluating the correctness of mathematical problems, and based on this method, we propose the Self-Critique method aimed at enhancing the mathematical capabilities of language models without the need for external supervisory models and manual annotations. Our experiments were conducted in both English and Chinese, and a 32-billion parameter model achieved state-of-the-art results among open-source language models on multiple datasets. Additionally, it surpassed several renowned proprietary models, including GPT-4-0613, on our proposed \textsc{MathUserEval} test set. Our method was applied during the development process of GLM-4 as a component to improve mathematical capabilities, achieving the best results on datasets such as \textsc{MathUserEval}, ape210k, GSM8k, and the Hungarian test, except GPT-4-Turbo.

\clearpage

\bibliographystyle{abbrv}
\bibliography{ref}

\begin{thebibliography}{10}

\bibitem{Claude}
Anthropic.
\newblock Introducing claude, 2023.

\bibitem{askell2021general}
A.~Askell, Y.~Bai, A.~Chen, D.~Drain, D.~Ganguli, T.~Henighan, A.~Jones, N.~Joseph, B.~Mann, N.~DasSarma, N.~Elhage, Z.~Hatfield-Dodds, D.~Hernandez, J.~Kernion, K.~Ndousse, C.~Olsson, D.~Amodei, T.~Brown, J.~Clark, S.~McCandlish, C.~Olah, and J.~Kaplan.
\newblock A general language assistant as a laboratory for alignment, 2021.

\bibitem{bai2023qwen}
J.~Bai, S.~Bai, et~al.
\newblock Qwen technical report, 2023.

\bibitem{bai2022constitutional}
Y.~Bai, S.~Kadavath, S.~Kundu, A.~Askell, J.~Kernion, A.~Jones, A.~Chen, A.~Goldie, A.~Mirhoseini, C.~McKinnon, C.~Chen, C.~Olsson, C.~Olah, D.~Hernandez, D.~Drain, D.~Ganguli, D.~Li, E.~Tran-Johnson, E.~Perez, J.~Kerr, J.~Mueller, J.~Ladish, J.~Landau, K.~Ndousse, K.~Lukosuite, L.~Lovitt, M.~Sellitto, N.~Elhage, N.~Schiefer, N.~Mercado, N.~DasSarma, R.~Lasenby, R.~Larson, S.~Ringer, S.~Johnston, S.~Kravec, S.~E. Showk, S.~Fort, T.~Lanham, T.~Telleen-Lawton, T.~Conerly, T.~Henighan, T.~Hume, S.~R. Bowman, Z.~Hatfield-Dodds, B.~Mann, D.~Amodei, N.~Joseph, S.~McCandlish, T.~Brown, and J.~Kaplan.
\newblock Constitutional ai: Harmlessness from ai feedback, 2022.

\bibitem{bai2023longbench}
Y.~Bai, X.~Lv, J.~Zhang, H.~Lyu, J.~Tang, Z.~Huang, Z.~Du, X.~Liu, A.~Zeng, L.~Hou, et~al.
\newblock Longbench: A bilingual, multitask benchmark for long context understanding.
\newblock {\em arXiv preprint arXiv:2308.14508}, 2023.

\bibitem{besta2023graph}
M.~Besta, N.~Blach, A.~Kubicek, R.~Gerstenberger, L.~Gianinazzi, J.~Gajda, T.~Lehmann, M.~Podstawski, H.~Niewiadomski, P.~Nyczyk, et~al.
\newblock Graph of thoughts: Solving elaborate problems with large language models.
\newblock {\em arXiv preprint arXiv:2308.09687}, 2023.

\bibitem{bras_Gao_Choi_2020}
Y.~Bisk, R.~Zellers, R.~Le~bras, J.~Gao, and Y.~Choi.
\newblock Piqa: Reasoning about physical commonsense in natural language.
\newblock {\em Proceedings of the AAAI Conference on Artificial Intelligence}, 34(05):7432--7439, Apr. 2020.

\bibitem{brown2020language}
T.~Brown, B.~Mann, N.~Ryder, M.~Subbiah, J.~D. Kaplan, P.~Dhariwal, A.~Neelakantan, P.~Shyam, G.~Sastry, A.~Askell, et~al.
\newblock Language models are few-shot learners.
\newblock {\em Advances in neural information processing systems}, 33:1877--1901, 2020.

\bibitem{abel-math}
E.~Chern, H.~Zou, X.~Li, J.~Hu, K.~Feng, J.~Li, and P.~Liu.
\newblock Generative ai for math: Abel.
\newblock \url{https://github.com/GAIR-NLP/abel}, 2023.

\bibitem{chowdhery2022palm}
A.~Chowdhery, S.~Narang, J.~Devlin, M.~Bosma, G.~Mishra, A.~Roberts, P.~Barham, H.~W. Chung, C.~Sutton, S.~Gehrmann, et~al.
\newblock Palm: Scaling language modeling with pathways.
\newblock {\em arXiv preprint arXiv:2204.02311}, 2022.

\bibitem{cobbe2021training}
K.~Cobbe, V.~Kosaraju, M.~Bavarian, M.~Chen, H.~Jun, L.~Kaiser, M.~Plappert, J.~Tworek, J.~Hilton, R.~Nakano, C.~Hesse, and J.~Schulman.
\newblock Training verifiers to solve math word problems, 2021.

\bibitem{deepseekai2024deepseek}
DeepSeek-AI, :, X.~Bi, D.~Chen, G.~Chen, S.~Chen, D.~Dai, C.~Deng, H.~Ding, K.~Dong, Q.~Du, Z.~Fu, H.~Gao, K.~Gao, W.~Gao, R.~Ge, K.~Guan, D.~Guo, J.~Guo, G.~Hao, Z.~Hao, Y.~He, W.~Hu, P.~Huang, E.~Li, G.~Li, J.~Li, Y.~Li, Y.~K. Li, W.~Liang, F.~Lin, A.~X. Liu, B.~Liu, W.~Liu, X.~Liu, X.~Liu, Y.~Liu, H.~Lu, S.~Lu, F.~Luo, S.~Ma, X.~Nie, T.~Pei, Y.~Piao, J.~Qiu, H.~Qu, T.~Ren, Z.~Ren, C.~Ruan, Z.~Sha, Z.~Shao, J.~Song, X.~Su, J.~Sun, Y.~Sun, M.~Tang, B.~Wang, P.~Wang, S.~Wang, Y.~Wang, Y.~Wang, T.~Wu, Y.~Wu, X.~Xie, Z.~Xie, Z.~Xie, Y.~Xiong, H.~Xu, R.~X. Xu, Y.~Xu, D.~Yang, Y.~You, S.~Yu, X.~Yu, B.~Zhang, H.~Zhang, L.~Zhang, L.~Zhang, M.~Zhang, M.~Zhang, W.~Zhang, Y.~Zhang, C.~Zhao, Y.~Zhao, S.~Zhou, S.~Zhou, Q.~Zhu, and Y.~Zou.
\newblock Deepseek llm: Scaling open-source language models with longtermism, 2024.

\bibitem{du2022glm}
Z.~Du, Y.~Qian, X.~Liu, M.~Ding, J.~Qiu, Z.~Yang, and J.~Tang.
\newblock Glm: General language model pretraining with autoregressive blank infilling.
\newblock In {\em Proceedings of the 60th Annual Meeting of the Association for Computational Linguistics (Volume 1: Long Papers)}, pages 320--335, 2022.

\bibitem{fu2023chain}
Y.~Fu, L.~Ou, M.~Chen, Y.~Wan, H.~Peng, and T.~Khot.
\newblock Chain-of-thought hub: A continuous effort to measure large language models' reasoning performance.
\newblock {\em arXiv preprint arXiv:2305.17306}, 2023.

\bibitem{gou2023tora}
Z.~Gou, Z.~Shao, Y.~Gong, Y.~Shen, Y.~Yang, M.~Huang, N.~Duan, and W.~Chen.
\newblock Tora: A tool-integrated reasoning agent for mathematical problem solving, 2023.

\bibitem{hendrycks2020measuring}
D.~Hendrycks, C.~Burns, S.~Basart, A.~Zou, M.~Mazeika, D.~Song, and J.~Steinhardt.
\newblock Measuring massive multitask language understanding.
\newblock {\em Proceedings of the International Conference on Learning Representations (ICLR)}, 2021.

\bibitem{hendrycks2021measuring-math}
D.~Hendrycks, C.~Burns, S.~Kadavath, A.~Arora, S.~Basart, E.~Tang, D.~Song, and J.~Steinhardt.
\newblock Measuring mathematical problem solving with the math dataset.
\newblock {\em arXiv preprint arXiv:2103.03874}, 2021.

\bibitem{NIPS2015_afdec700}
K.~M. Hermann, T.~Kocisky, E.~Grefenstette, L.~Espeholt, W.~Kay, M.~Suleyman, and P.~Blunsom.
\newblock Teaching machines to read and comprehend.
\newblock In C.~Cortes, N.~Lawrence, D.~Lee, M.~Sugiyama, and R.~Garnett, editors, {\em Advances in Neural Information Processing Systems}, volume~28. Curran Associates, Inc., 2015.

\bibitem{hou2024chatglmrlhf}
Z.~Hou, Y.~Niu, Z.~Du, X.~Zhang, X.~Liu, A.~Zeng, Q.~Zheng, M.~Huang, H.~Wang, J.~Tang, and Y.~Dong.
\newblock Chatglm-rlhf: Practices of aligning large language models with human feedback, 2024.

\bibitem{kaplan2020scaling}
J.~Kaplan, S.~McCandlish, T.~Henighan, T.~B. Brown, B.~Chess, R.~Child, S.~Gray, A.~Radford, J.~Wu, and D.~Amodei.
\newblock Scaling laws for neural language models.
\newblock {\em arXiv preprint arXiv:2001.08361}, 2020.

\bibitem{ke2023critiquellm}
P.~Ke, B.~Wen, Z.~Feng, X.~Liu, X.~Lei, J.~Cheng, S.~Wang, A.~Zeng, Y.~Dong, H.~Wang, et~al.
\newblock Critiquellm: Scaling llm-as-critic for effective and explainable evaluation of large language model generation.
\newblock {\em arXiv preprint arXiv:2311.18702}, 2023.

\bibitem{koncel2016mawps}
R.~Koncel-Kedziorski, S.~Roy, A.~Amini, N.~Kushman, and H.~Hajishirzi.
\newblock Mawps: A math word problem repository.
\newblock In {\em Proceedings of the 2016 conference of the north american chapter of the association for computational linguistics: human language technologies}, pages 1152--1157, 2016.

\bibitem{kumar2021adversarial}
V.~Kumar, R.~Maheshwary, and V.~Pudi.
\newblock Adversarial examples for evaluating math word problem solvers.
\newblock In {\em Findings of the Association for Computational Linguistics: EMNLP 2021}, pages 2705--2712, 2021.

\bibitem{kwiatkowski2019natural}
T.~Kwiatkowski, J.~Palomaki, O.~Redfield, M.~Collins, A.~Parikh, C.~Alberti, D.~Epstein, I.~Polosukhin, J.~Devlin, K.~Lee, et~al.
\newblock Natural questions: a benchmark for question answering research.
\newblock {\em Transactions of the Association for Computational Linguistics}, 7:453--466, 2019.

\bibitem{lee2023rlaif}
H.~Lee, S.~Phatale, H.~Mansoor, T.~Mesnard, J.~Ferret, K.~Lu, C.~Bishop, E.~Hall, V.~Carbune, A.~Rastogi, and S.~Prakash.
\newblock Rlaif: Scaling reinforcement learning from human feedback with ai feedback, 2023.

\bibitem{li-etal-2022-csl}
Y.~Li, Y.~Zhang, Z.~Zhao, L.~Shen, W.~Liu, W.~Mao, and H.~Zhang.
\newblock {CSL}: A large-scale {C}hinese scientific literature dataset.
\newblock In {\em Proceedings of the 29th International Conference on Computational Linguistics}, pages 3917--3923. International Committee on Computational Linguistics, Oct. 2022.

\bibitem{lightman2023lets-verify}
H.~Lightman, V.~Kosaraju, Y.~Burda, H.~Edwards, B.~Baker, T.~Lee, J.~Leike, J.~Schulman, I.~Sutskever, and K.~Cobbe.
\newblock Let's verify step by step, 2023.

\bibitem{ling2017program}
W.~Ling, D.~Yogatama, C.~Dyer, and P.~Blunsom.
\newblock Program induction by rationale generation: Learning to solve and explain algebraic word problems.
\newblock In {\em Proceedings of the 55th Annual Meeting of the Association for Computational Linguistics (Volume 1: Long Papers)}, pages 158--167, 2017.

\bibitem{liu2023alignbench}
X.~Liu, X.~Lei, S.~Wang, Y.~Huang, Z.~Feng, B.~Wen, J.~Cheng, P.~Ke, Y.~Xu, W.~L. Tam, et~al.
\newblock Alignbench: Benchmarking chinese alignment of large language models.
\newblock {\em arXiv preprint arXiv:2311.18743}, 2023.

\bibitem{liu2023agentbench}
X.~Liu, H.~Yu, H.~Zhang, Y.~Xu, X.~Lei, H.~Lai, Y.~Gu, H.~Ding, K.~Men, K.~Yang, et~al.
\newblock Agentbench: Evaluating llms as agents.
\newblock {\em arXiv preprint arXiv:2308.03688}, 2023.

\bibitem{wizardmath}
H.~Luo, Q.~Sun, C.~Xu, P.~Zhao, J.~Lou, C.~Tao, X.~Geng, Q.~Lin, S.~Chen, and D.~Zhang.
\newblock Wizardmath: Empowering mathematical reasoning for large language models via reinforced evol-instruct.
\newblock {\em arXiv preprint arXiv:2308.09583}, 2023.

\bibitem{mishra2022numglue}
S.~Mishra, A.~Mitra, N.~Varshney, B.~Sachdeva, P.~Clark, C.~Baral, and A.~Kalyan.
\newblock Numglue: A suite of fundamental yet challenging mathematical reasoning tasks.
\newblock In {\em Proceedings of the 60th Annual Meeting of the Association for Computational Linguistics (Volume 1: Long Papers)}, pages 3505--3523, 2022.

\bibitem{xsum-emnlp}
S.~Narayan, S.~B. Cohen, and M.~Lapata.
\newblock Don't give me the details, just the summary! {T}opic-aware convolutional neural networks for extreme summarization.
\newblock In {\em Proceedings of the 2018 Conference on Empirical Methods in Natural Language Processing}, Brussels, Belgium, 2018.

\bibitem{openai2023gpt4}
OpenAI.
\newblock Gpt-4 technical report, 2023.

\bibitem{ouyang2022training}
L.~Ouyang, J.~Wu, X.~Jiang, D.~Almeida, C.~Wainwright, P.~Mishkin, C.~Zhang, S.~Agarwal, K.~Slama, A.~Ray, et~al.
\newblock Training language models to follow instructions with human feedback.
\newblock {\em Advances in Neural Information Processing Systems}, 35:27730--27744, 2022.

\bibitem{testing_language_models_on_a_held_out_high_school_national_finals_exam}
K.~Paster.
\newblock Testing language models on a held-out high school national finals exam.
\newblock \url{https://huggingface.co/datasets/keirp/hungarian_national_hs_finals_exam}, 2023.

\bibitem{polu2020generative}
S.~Polu and I.~Sutskever.
\newblock Generative language modeling for automated theorem proving.
\newblock {\em arXiv preprint arXiv:2009.03393}, 2020.

\bibitem{rafailov2023direct}
R.~Rafailov, A.~Sharma, E.~Mitchell, S.~Ermon, C.~D. Manning, and C.~Finn.
\newblock Direct preference optimization: Your language model is secretly a reward model, 2023.

\bibitem{saxton2019analysing}
D.~Saxton, E.~Grefenstette, F.~Hill, and P.~Kohli.
\newblock Analysing mathematical reasoning abilities of neural models.
\newblock {\em arXiv preprint arXiv:1904.01557}, 2019.

\bibitem{scao2022bloom}
T.~L. Scao, A.~Fan, C.~Akiki, E.~Pavlick, S.~Ili{\'c}, D.~Hesslow, R.~Castagn{\'e}, A.~S. Luccioni, F.~Yvon, M.~Gall{\'e}, et~al.
\newblock Bloom: A 176b-parameter open-access multilingual language model.
\newblock {\em arXiv preprint arXiv:2211.05100}, 2022.

\bibitem{shao2023characterllm}
Y.~Shao, L.~Li, J.~Dai, and X.~Qiu.
\newblock Character-llm: A trainable agent for role-playing, 2023.

\bibitem{suzgun2022challenging}
M.~Suzgun, N.~Scales, N.~Sch{\"a}rli, S.~Gehrmann, Y.~Tay, H.~W. Chung, A.~Chowdhery, Q.~V. Le, E.~H. Chi, D.~Zhou, et~al.
\newblock Challenging big-bench tasks and whether chain-of-thought can solve them.
\newblock {\em arXiv preprint arXiv:2210.09261}, 2022.

\bibitem{2023internlm}
I.~Team.
\newblock Internlm: A multilingual language model with progressively enhanced capabilities.
\newblock \url{https://github.com/InternLM/InternLM}, 2023.

\bibitem{touvron2023llama}
H.~Touvron, T.~Lavril, G.~Izacard, X.~Martinet, M.-A. Lachaux, T.~Lacroix, B.~Rozi{\`e}re, N.~Goyal, E.~Hambro, F.~Azhar, et~al.
\newblock Llama: Open and efficient foundation language models.
\newblock {\em arXiv preprint arXiv:2302.13971}, 2023.

\bibitem{touvron2023llama2}
H.~Touvron, L.~Martin, K.~Stone, P.~Albert, A.~Almahairi, Y.~Babaei, N.~Bashlykov, S.~Batra, P.~Bhargava, S.~Bhosale, D.~Bikel, L.~Blecher, C.~C. Ferrer, M.~Chen, G.~Cucurull, D.~Esiobu, J.~Fernandes, J.~Fu, W.~Fu, B.~Fuller, C.~Gao, V.~Goswami, N.~Goyal, A.~Hartshorn, S.~Hosseini, R.~Hou, H.~Inan, M.~Kardas, V.~Kerkez, M.~Khabsa, I.~Kloumann, A.~Korenev, P.~S. Koura, M.-A. Lachaux, T.~Lavril, J.~Lee, D.~Liskovich, Y.~Lu, Y.~Mao, X.~Martinet, T.~Mihaylov, P.~Mishra, I.~Molybog, Y.~Nie, A.~Poulton, J.~Reizenstein, R.~Rungta, K.~Saladi, A.~Schelten, R.~Silva, E.~M. Smith, R.~Subramanian, X.~E. Tan, B.~Tang, R.~Taylor, A.~Williams, J.~X. Kuan, P.~Xu, Z.~Yan, I.~Zarov, Y.~Zhang, A.~Fan, M.~Kambadur, S.~Narang, A.~Rodriguez, R.~Stojnic, S.~Edunov, and T.~Scialom.
\newblock Llama 2: Open foundation and fine-tuned chat models, 2023.

\bibitem{tu2024charactereval}
Q.~Tu, S.~Fan, Z.~Tian, and R.~Yan.
\newblock Charactereval: A chinese benchmark for role-playing conversational agent evaluation, 2024.

\bibitem{volske-etal-2017-tl}
M.~V{\"o}lske, M.~Potthast, S.~Syed, and B.~Stein.
\newblock {TL};{DR}: Mining {R}eddit to learn automatic summarization.
\newblock In L.~Wang, J.~C.~K. Cheung, G.~Carenini, and F.~Liu, editors, {\em Proceedings of the Workshop on New Frontiers in Summarization}, pages 59--63, Copenhagen, Denmark, Sept. 2017. Association for Computational Linguistics.

\bibitem{wang2023mathshepherd}
P.~Wang, L.~Li, Z.~Shao, R.~X. Xu, D.~Dai, Y.~Li, D.~Chen, Y.~Wu, and Z.~Sui.
\newblock Math-shepherd: Verify and reinforce llms step-by-step without human annotations, 2023.

\bibitem{wang2023selfconsistency}
X.~Wang, J.~Wei, D.~Schuurmans, Q.~Le, E.~Chi, S.~Narang, A.~Chowdhery, and D.~Zhou.
\newblock Self-consistency improves chain of thought reasoning in language models, 2023.

\bibitem{wang-etal-2017-deep-math23}
Y.~Wang, X.~Liu, and S.~Shi.
\newblock Deep neural solver for math word problems.
\newblock In M.~Palmer, R.~Hwa, and S.~Riedel, editors, {\em Proceedings of the 2017 Conference on Empirical Methods in Natural Language Processing}, pages 845--854, Copenhagen, Denmark, Sept. 2017. Association for Computational Linguistics.

\bibitem{wei2023chainofthought}
J.~Wei, X.~Wang, D.~Schuurmans, M.~Bosma, B.~Ichter, F.~Xia, E.~Chi, Q.~Le, and D.~Zhou.
\newblock Chain-of-thought prompting elicits reasoning in large language models, 2023.

\bibitem{wei2023cmath}
T.~Wei, J.~Luan, W.~Liu, S.~Dong, and B.~Wang.
\newblock Cmath: Can your language model pass chinese elementary school math test?, 2023.

\bibitem{yang2023large}
C.~Yang, X.~Wang, Y.~Lu, H.~Liu, Q.~V. Le, D.~Zhou, and X.~Chen.
\newblock Large language models as optimizers.
\newblock {\em arXiv preprint arXiv:2309.03409}, 2023.

\bibitem{yang2023skymath}
L.~Yang, H.~Yang, W.~Cheng, L.~Lin, C.~Li, Y.~Chen, L.~Liu, J.~Pan, T.~Wei, B.~Li, L.~Zhao, L.~Wang, B.~Zhu, G.~Li, X.~Wu, X.~Luo, and R.~Hu.
\newblock Skymath: Technical report, 2023.

\bibitem{yao2023tree}
S.~Yao, D.~Yu, J.~Zhao, I.~Shafran, T.~L. Griffiths, Y.~Cao, and K.~Narasimhan.
\newblock Tree of thoughts: Deliberate problem solving with large language models.
\newblock {\em arXiv preprint arXiv:2305.10601}, 2023.

\bibitem{Yi}
Yi.
\newblock A series of large language models trained from scratch by developers at 01-ai.
\newblock \url{https://github.com/01-ai/Yi}, 2023.

\bibitem{metamath}
L.~Yu, W.~Jiang, H.~Shi, J.~Yu, Z.~Liu, Y.~Zhang, J.~T. Kwok, Z.~Li, A.~Weller, and W.~Liu.
\newblock Metamath: Bootstrap your own mathematical questions for large language models.
\newblock {\em arXiv preprint arXiv:2309.12284}, 2023.

\bibitem{yuan2023scaling-mathrft}
Z.~Yuan, H.~Yuan, C.~Li, G.~Dong, K.~Lu, C.~Tan, C.~Zhou, and J.~Zhou.
\newblock Scaling relationship on learning mathematical reasoning with large language models, 2023.

\bibitem{yuan2023large-math401-llm}
Z.~Yuan, H.~Yuan, C.~Tan, W.~Wang, and S.~Huang.
\newblock How well do large language models perform in arithmetic tasks?, 2023.

\bibitem{yue2023mammoth}
X.~Yue, X.~Qu, G.~Zhang, Y.~Fu, W.~Huang, H.~Sun, Y.~Su, and W.~Chen.
\newblock Mammoth: Building math generalist models through hybrid instruction tuning.
\newblock {\em arXiv preprint arXiv:2309.05653}, 2023.

\bibitem{zeng2022glm}
A.~Zeng, X.~Liu, Z.~Du, Z.~Wang, H.~Lai, M.~Ding, Z.~Yang, Y.~Xu, W.~Zheng, X.~Xia, et~al.
\newblock Glm-130b: An open bilingual pre-trained model.
\newblock {\em arXiv preprint arXiv:2210.02414}, 2022.

\bibitem{zhang2024sciglm}
D.~Zhang, Z.~Hu, S.~Zhoubian, Z.~Du, K.~Yang, Z.~Wang, Y.~Yue, Y.~Dong, and J.~Tang.
\newblock Sciglm: Training scientific language models with self-reflective instruction annotation and tuning, 2024.

\bibitem{Zhang2023EvaluatingTP-gaokaobench}
X.~Zhang, C.~Li, Y.~Zong, Z.~Ying, L.~He, and X.~Qiu.
\newblock Evaluating the performance of large language models on gaokao benchmark.
\newblock 2023.

\bibitem{zhao2020ape210k}
W.~Zhao, M.~Shang, Y.~Liu, L.~Wang, and J.~Liu.
\newblock Ape210k: A large-scale and template-rich dataset of math word problems, 2020.

\bibitem{zheng2023judging}
L.~Zheng, W.-L. Chiang, Y.~Sheng, S.~Zhuang, Z.~Wu, Y.~Zhuang, Z.~Lin, Z.~Li, D.~Li, E.~P. Xing, H.~Zhang, J.~E. Gonzalez, and I.~Stoica.
\newblock Judging llm-as-a-judge with mt-bench and chatbot arena, 2023.

\bibitem{zhong2023agieval}
W.~Zhong, R.~Cui, Y.~Guo, Y.~Liang, S.~Lu, Y.~Wang, A.~Saied, W.~Chen, and N.~Duan.
\newblock Agieval: A human-centric benchmark for evaluating foundation models, 2023.

\bibitem{zhou2023characterglm}
J.~Zhou, Z.~Chen, D.~Wan, B.~Wen, Y.~Song, J.~Yu, Y.~Huang, L.~Peng, J.~Yang, X.~Xiao, et~al.
\newblock Characterglm: Customizing chinese conversational ai characters with large language models.
\newblock {\em arXiv preprint arXiv:2311.16832}, 2023.

\bibitem{zhou2023mathattack}
Z.~Zhou, Q.~Wang, M.~Jin, J.~Yao, J.~Ye, W.~Liu, W.~Wang, X.~Huang, and K.~Huang.
\newblock Mathattack: Attacking large language models towards math solving ability.
\newblock {\em arXiv preprint arXiv:2309.01686}, 2023.

\end{thebibliography}

\appendix

\section{Case Study}
\subsection{Case Study of Math-Critique}

\begin{table*}[]
\caption{Math-Critique Example (1)}
\label{Math-Critique Example (1)}
\centering
\renewcommand\arraystretch{1.8}
\resizebox{\textwidth}{!}{

\begin{tabular}{p{0.13\textwidth}p{0.87\textwidth}}
\hline
Question & \begin{CJK}{UTF8}{gbsn}1/x+x三次方的原函数是多少。\end{CJK} 

\\ \hdashline

Question  (Translate) & What is the original function of 1/x+x cubed.
\\
\hline

Reference & \begin{CJK}{UTF8}{gbsn}首先，我们可以将 $1/x+x^3$ 重新写成 $x^{-1}+x^3$，然后利用幂函数和多项式函数的求导原理求出它的原函数。 具体来说，我们得到：$$\begin{aligned} \int (x^{-1}+x^3)\mathrm{d}x &= \int x^{-1}\mathrm{d}x + \int x^3\mathrm{d}x = \ln|x|+\frac{1}{4}x^4+C, \end{aligned}$$其中 $C$ 是不定积分常数。 因此，$1/x+x^3$ 的原函数是 $\ln|x|+\frac{1}{4}x^4+C$。\end{CJK}
\\ \hdashline
Reference  (Translate) & 
First, we can rewrite $1/x+x^3$ as $x^{-1}+x^3$, and then use the derivation principles of power functions and polynomial functions to find its original function. Specifically, we get: $$\begin{aligned} \int (x^{-1}+x^3)\mathrm{d}x &= \int x^{-1}\mathrm{d}x + \int x^3\mathrm{d}x = \ln|x|+\frac{1}{4}x^4+C, \end{aligned}$$where $C$ is the indefinite integral constant. Therefore, the original function of $1/x+x^3$ is $\ln|x|+\frac{1}{4}x^4+C$.

\\ \hline
Model Answer by GLM4 & \begin{CJK}{UTF8}{gbsn}要求函数 \( f(x) = \frac{1}{x} + x^3 \) 的原函数（不定积分），我们需要分别对这两项进行积分。  对于 \( \frac{1}{x} \)，其原函数是 \( \ln{|x|} \)（自然对数）。对于 \( x^3 \)，我们可以利用幂函数的积分规则，即对于 \( x^n \)（其中 \( n eq -1 \)），其原函数是 \( \frac{x^{n+1}}{n+1} \)。因此，\( x^3 \) 的原函数是 \( \frac{x^4}{4} \)。将这两项的原函数相加，我们得到 \( f(x) \) 的原函数：\[ \int \left( \frac{1}{x} + x^3 \right) dx = \ln{|x|} + \frac{x^4}{4} + C \]其中 \( C \) 是积分常数。这个结果表示在定义域内（除了 \( x = 0 \) 的情况，因为 \( \frac{1}{x} \) 在 \( x = 0 \) 处未定义），\( f(x) \) 的一个原函数。由于 \( \ln{|x|} \) 在 \( x = 0 \) 处未定义，所以这个原函数在 \( x = 0 \) 处也是不连续的。\end{CJK}
\\ \hdashline
Model Answer (Translate) & 
To find the primitive function (indefinite integral) of the function \( f(x) = \frac{1}{x} + x^3 \), we need to integrate these two terms separately. For \( \frac{1}{x} \), the original function is \( \ln{|x|} \) (natural logarithm). For \( x^3 \), we can use the integration rule of the power function, that is, for \( x^n \) (where \( n eq -1 \)), the original function is\( \frac{x^{n+1}}{n+1} \). Therefore, the original function of \( x^3 \) is \( \frac{x^4}{4} \). Adding the original functions of these two terms, we get the original function of \( f(x) \): \[ \int \left( \frac{1}{x} + x^ 3 \right) dx = \ln{|x|} + \frac{x^4}{4} + C \]where \( C \) is the integration constant. This result means that within the domain (except for the case of \( x = 0 \), since \( \frac{1}{x} \) is undefined at \( x = 0 \)), \( f(x) A primitive function of \). Since \( \ln{|x|} \) is undefined at \( x = 0 \), the original function is also discontinuous at \( x = 0 \).
\\ \hline
Critique Result & \begin{CJK}{UTF8}{gbsn}AI助手的回答非常准确和详细，它正确地解释了如何求解给定函数的原函数。AI助手首先分别对函数的两部分进行积分，然后将它们的结果相加，得到了原函数。此外，AI助手还注意到了在x=0处，原函数是不连续的，这是一个重要的观察。总的来说，AI助手的回答与参考答案非常接近，没有明显的缺点。因此，我会给它一个高分。评分：[[9]\end{CJK}
\\ \hdashline
Critique Result (Translate) & 
The AI assistant's answer is very accurate and detailed, and it correctly explains how to solve the original function of a given function. The AI assistant first integrates the two parts of the function separately and then adds their results to obtain the original function. In addition, the AI assistant also noticed that at x=0, the original function is discontinuous, which is an important observation. Overall, the AI assistant's answers are very close to the reference answers, with no obvious shortcomings. Therefore, I would give it a high score. Rating:[[9]

\\ \hline

\end{tabular}

}
\end{table*}

\begin{table*}[]
\footnotesize
\setlength{\tabcolsep}{15pt}
\caption{Math-Critique Example (2)}
\label{Math-Critique Example (2)}
\renewcommand\arraystretch{2}
\noindent\makebox[\textwidth]{
\resizebox{\textwidth}{!}{

\begin{tabular}{p{0.1\textwidth}p{0.9\textwidth}}
\hline
Question & \begin{CJK}{UTF8}{gbsn}三阶矩阵A有三个特征值0,3,5，对应的特征向量分别为u,v,w，则线性方程组Ax = v + w的通解为？\end{CJK} 

\\ \hdashline

Question  (Translate) & The third-order matrix A has three eigenvalues 0, 3, and 5, and the corresponding eigenvectors are u, v, and w respectively. What is the general solution to the linear equation system Ax = v + w?
\\
\hline

Reference & \begin{CJK}{UTF8}{gbsn}给定矩阵A有特征值0,3,5，对应的特征向量为u,v,w。线性方程组Ax = v + w，我们可以写作Au = 0 * u，Av = 3 * v，和Aw = 5 * w。所以A(1/3*v)=v,A(1/5*w)=w，左右两边相加得A(1/3*v+1/5*w)= v + w。又因为Ax = v + w，所以1/3 * v + 1/5 * w是Ax = v + w的一个特解，因为Au = 0，所以u是Au = 0的基础解系，因此方程Ax = v + w的通解是x =ku+  1/3 * v + 1/5 * w。
\end{CJK}
\\ \hdashline
Reference  (Translate) & A given matrix A has eigenvalues 0, 3, and 5, and the corresponding eigenvectors are u, v, and w. For the linear system of equations Ax = v + w, we can write Au = 0 * u, Av = 3 * v, and Aw = 5 * w. So A(1/3*v)=v, A(1/5*w)=w, adding the left and right sides gives A(1/3*v+1/5*w)= v + w. And because Ax = v + w, 1/3 * v + 1/5 * w is a special solution of Ax = v + w. Because Au = 0, so u is the basic solution system of Au = 0, so the equation Ax The general solution of = v + w is x =ku+ 1/3 * v + 1/5 * w.

\\ \hline
Model Answer by ChatGLM3-32B-Math & \begin{CJK}{UTF8}{gbsn}给定三阶矩阵 \( A \) 的三个特征值分别为 \( 0, 3, 5 \)，对应的特征向量分别为 \( u, v, w \)。要找线性方程组 \( Ax = v + w \) 的通解，我们可以利用特征值和特征向量的性质。  首先，我们知道 \( Av = 3v \) 和 \( Aw = 5w \)。因此，\( v \) 和 \( w \) 分别是特征值 \( 3 \) 和 \( 5 \) 对应的特征向量。  线性方程组 \( Ax = v + w \) 可以重写为 \( Ax - v - w = 0 \)。  我们可以将 \( x \) 写成 \( x = pu + qv + rw \)，其中 \( p, q, r \) 是待定系数。  将 \( x \) 代入方程 \( Ax - v - w = 0 \)，我们得到：  \[ A(pu + qv + rw) - v - w = 0 \] ... 由于 \( v \) 和 \( w \) 是线性无关的特征向量（因为它们对应不同的特征值），上式成立的唯一可能是 \( 3q - 1 = 0 \) 和 \( 5r - 1 = 0 \)。  解这两个方程，我们得到 \( q = \frac{1}{3} \) 和 \( r = \frac{1}{5} \)。  因此，方程 \( Ax = v + w \) 的通解为：  \[ x = pu + \frac{1}{3}v + \frac{1}{5}w \]  其中 \( p \) 是任意实数。  所以，通解的形式是：  \[ x = p\mathbf{u} + \frac{1}{3}\mathbf{v} + \frac{1}{5}\mathbf{w} \]  这里的 \( \mathbf{u} \) 是特征值 \( 0 \) 对应的特征向量，\( \mathbf{v} \) 和 \( \mathbf{w} \) 分别是特征值 \( 3 \) 和 \( 5 \) 对应的特征向量。\( p \) 是任意实数，可以取任何值。
\end{CJK}
\\ \hdashline
Model Answer (Translate) & 
Given the three eigenvalues of the third-order matrix \( A \) are \( 0, 3, 5 \), the corresponding eigenvectors are \( u, v, w \). To find the general solution to a system of linear equations \( Ax = v + w \), we can exploit the properties of eigenvalues and eigenvectors.    First of all, we know \( Av = 3v \) and \( Aw = 5w \). Therefore, \( v \) and \( w \) are the eigenvectors corresponding to the eigenvalues \( 3 \) and \( 5 \) respectively.    The system of linear equations \( Ax = v + w \) can be rewritten as \( Ax - v - w = 0 \).    We can write \( x \) as \( x = pu + qv + rw \), where \( p, q, r \) are the undetermined coefficients.    Substituting \( x \) into the equation\( Ax - v - w = 0 \), we get:  \[ A(pu + qv + rw) - v - w = 0 \] ...    Therefore, the general solution of equation \( Ax = v + w \) is:  \[ x = pu + \frac{1}{3}v + \frac{1}{5}w \]   Where \( p \) is any real number.    So, the general solution is in the form:  \[ x = p\mathbf{u} + \frac{1}{3}\mathbf{v} + \frac{1}{5}\mathbf {w} \]   Here\( \mathbf{u} \) is the eigenvector corresponding to the eigenvalue \( 0 \), \( \mathbf{v} \) and \( \mathbf{w} \) are the eigenvectors corresponding to the eigenvalues \( 3 \) and \( 5 \) respectively. \( p \) is any real number and can take on any value.
\\ \hline
Critique Result & \begin{CJK}{UTF8}{gbsn}AI助手的回答非常详细，它首先解释了如何利用特征值和特征向量的性质来解决这个问题，然后逐步进行了详细的步骤，包括将方程重写，将x写成待定系数的形式，然后将这些系数代入方程，最后解出这些系数。这个答案不仅准确，而且深入，提供了详细的解释和步骤，使得读者能够理解如何解决这个问题。与参考答案相比，AI助手的答案更加详细和深入，因此我认为AI助手的回答质量非常高。评分：[[9]]。\end{CJK}
\\ \hdashline
Critique Result (Translate) & 
The AI assistant's answer is very detailed. It first explains how to use the properties of eigenvalues and eigenvectors to solve this problem, and then goes through the detailed steps, including rewriting the equation, writing x in the form of undetermined coefficients, and then converting these The coefficients are substituted into the equation and finally the coefficients are solved. This answer is not only accurate but also in-depth, providing detailed explanations and steps so the reader can understand how to solve the problem. Compared with the reference answers, the AI assistant's answers are more detailed and in-depth, so I think the quality of the AI assistant's answers is very high. Rating: [[9]].

\\ \hline

\end{tabular}

}
}
\end{table*}

\begin{table*}[]
\caption{Math-Critique Example (3)}
\label{Math-Critique Example (3)}
\centering
	\renewcommand\arraystretch{2}
\resizebox{\textwidth}{!}{

\begin{tabular}{p{0.13\textwidth}p{0.87\textwidth}}
\hline
Question & \begin{CJK}{UTF8}{gbsn}二元一次方程4x-3y=12,当x=0,1,2,3时,y=\end{CJK} 

\\ \hdashline

Question  (Translate) & Quadratic equation 4x-3y=12, when x=0,1,2,3, y=
\\
\hline

Reference & \begin{CJK}{UTF8}{gbsn}将二元一次方程4x-3y=12改写为y的表达式： 4x-3y=12 -3y = -4x + 12 y = (4/3)x - 4   当x=0时，y=(-4)； 当x=1时，y=(4/3)-4=-8/3； 当x=2时，y=(8/3)-4=-4/3； 当x=3时，y=(4)-4=0。  因此，当x分别为0、1、2、3时，y的值分别为-4、-8/3、-4/3和0。
\end{CJK}
\\ \hdashline
Reference  (Translate) & Rewrite the linear equation 4x-3y=12 into the expression of y: 4x-3y=12 -3y = -4x + 12 y = (4/3)x - 4   When x= When 0, y=(-4); When x=1, y=(4/3)-4=-8/3; When x=2, y=(8/3)-4 =-4/3; When x=3, y=(4)-4=0.   Therefore, when x is 0, 1, 2, and 3 respectively, the values of y are -4, -8/3, -4/3, and 0 respectively.

\\ \hline
Model Answer by Claude-2 & \begin{CJK}{UTF8}{gbsn}
    * 给定二元一次方程:4x - 3y = 12 * 将x替换为给定的数值,可以得到y的值: * x = 0 代入方程,得到: -3y = 12, y = -4 * x = 1 代入方程,得到: 4 - 3y = 12, -3y = 8, y = -2  * x = 2 代入方程,得到: 8 - 3y = 12, -3y = 4, y = 0 * x = 3 代入方程,得到:12 - 3y = 12, -3y = 0, y = 2 * 综上,当x=0,1,2,3时,y依次等于-4,-2,0,2。  所以应该填写:-4,-2,0,2。
\end{CJK}
\\ \hdashline
Model Answer (Translate) & 
* Given a linear equation of two variables: 4x - 3y = 12 * Replace x with the given value, you can get the value of y: * x = 0 Substitute into the equation, you get: -3y = 12, y = - 4 * x = 1 Substituting into the equation, we get: 4 - 3y = 12, -3y = 8, y = -2  * x = 2 Substituting into the equation, we get: 8 - 3y = 12, -3y = 4, y = 0 * x = 3 Substituting into the equation, we get: 12 - 3y = 12, -3y = 0, y = 2 * In summary, when x=0,1,2,3, y is equal to -4 in turn ,-2,0,2.   So you should fill in:-4,-2,0,2.
\\ \hline
Critique Result & \begin{CJK}{UTF8}{gbsn}AI助手的回答在计算过程中出现了错误。当x=1时，AI助手错误地计算出y=-2，而实际上应该是y=-8/3。同样，当x=3时，AI助手错误地计算出y=2，而实际上应该是y=0。因此，AI助手的回答在准确性和深度上都存在问题。然而，AI助手的回答在结构和详细程度上是正确的，它详细地解释了如何计算y的值，并给出了每个x值对应的y值。总的来说，AI助手的回答虽然有一些错误，但是仍然提供了一些有用的信息。因此，我会给AI助手的回答打分为3。  评分：[[3]]\end{CJK}
\\ \hdashline
Critique Result (Translate) & 
There was an error in the calculation process of the AI assistant's answer. When x=1, the AI assistant incorrectly calculates y=-2, when it should actually be y=-8/3. Likewise, when x=3, the AI assistant incorrectly calculates y=2, when it should actually be y=0. Therefore, the AI assistant's answers have problems with accuracy and depth. However, the AI assistant's answer is correct in structure and level of detail. It explains in detail how to calculate the value of y and gives the corresponding y value for each x value. Overall, although the AI assistant's answers contained some errors, they still provided some useful information. Therefore, I would rate the AI assistant's answer a 3.   Rating: [[3]]

\\ \hline

\end{tabular}

}
\end{table*}
We have provided several examples of scoring by Math-Critique. In the examples from Table~\ref{Math-Critique Example (1)} and Table~\ref{Math-Critique Example (2)}, we observed that the model provided correct answers. However, the answer formats differed from the standard answers, specifically regarding fraction expression and the selection of unknown variables. These differences are equivalent, yet traditional evaluation methods struggle to judge them accurately. However, Math-Critique correctly scored these two examples and provided reasonable evaluations.

In the example from Table~\ref{Math-Critique Example (3)}, the model made a mistake in the calculation process. Math-Critique accurately pinpointed the error location, and since the model correctly solved a part of the problem, Math-Critique awarded a score of 3 points.
\subsection{Case Study of Mathematical Models}
\begin{table*}[]
\footnotesize
\setlength{\tabcolsep}{15pt}
\caption{ChatGLM3-32B-Math Example (1)}
\label{Math Example (1)}
\centering
	\renewcommand\arraystretch{2}
  \noindent\makebox[\textwidth]{
\resizebox{\textwidth}{!}{

\begin{tabular}{p{0.1\textwidth}p{0.9\textwidth}}

\hline
\textbf{Question} & \begin{CJK}{UTF8}{gbsn}已知一个两位数的十位数字比个位数字大 2，两位数字的积比这个两位数小34，求这个两位数。\end{CJK} 
\\ \hdashline
\textbf{Question (Translate)} & Given a two-digit number where the tens digit is 2 greater than the units digit, and the product of the digits is 34 less than the number itself, find this two-digit number. \\
\hline
\textbf{Reference} & \begin{CJK}{UTF8}{gbsn}设这个两位数为10x+y（x表示十位数，y表示个位数），根据题意可得：（1）x = y + 2 （十位数比个位数大2）（2）xy = 10x+y - 34 （两位数字的积比这个两位数小34）将（1）式代入（2）式得到：(y+2)y = 10(y+2)+y-34整理得到  \(y^2 + 2y = 11y + 20 - 34\)，整理得 \(y^2 - 9y + 14 = 0\)，解方程得到 y=7 或 y=2.所以x=7+2=9或x=2+2=4所以这个数为97或者42.答案：97或42。\end{CJK} 
\\ \hdashline
\textbf{Reference (Translate)} & Let the two-digit number be \(10x+y\) (where \(x\) is the tens digit and \(y\) is the units digit). According to the problem: (1) \(x = y + 2\) (the tens digit is 2 greater than the units digit) (2) \(xy = 10x + y - 34\) (the product of the digits is 34 less than the number itself). Substituting (1) into (2) gives: \((y+2)y = 10(y+2) + y - 34\), which simplifies to \(y^2 + 2y = 11y + 20 - 34\), and further simplifies to \(y^2 - 9y + 14 = 0\). Solving the equation yields \(y = 7\) or \(y = 2\). Thus, \(x = 7 + 2 = 9\) or \(x = 2 + 2 = 4\), making the number 97 or 42. Answer: 97 or 42. \\
\hline
\textbf{GPT-4-0613} & \begin{CJK}{UTF8}{gbsn}设这个两位数为10x+y，其中x为十位数，y为个位数。根据题意，我们有以下两个方程：x = y + 2 （十位数字比个位数字大2）xy = 10x + y - 34 （两位数字的积比这个两位数小34）将第一个方程代入第二个方程，我们得到：(y + 2)y = 10(y + 2) + y - 34解这个方程，我们得到y = 4。将y = 4代入第一个方程，我们得到x = 6。所以，这个两位数是64。\end{CJK} 
\\ \hdashline
\textbf{GPT-4-0613 (Translate)} & Let the two-digit number be \(10x+y\), where \(x\) is the tens digit and \(y\) is the units digit. According to the problem, we have two equations: \(x = y + 2\) (the tens digit is 2 greater than the units digit) and \(xy = 10x + y - 34\) (the product of the digits is 34 less than the number itself). Substituting the first equation into the second, we get: \((y + 2)y = 10(y + 2) + y - 34\). Solving this equation yields \(y = 4\). Substituting \(y = 4\) back into the first equation gives \(x = 6\). Therefore, the two-digit number is 64. \\
\hline
\textbf{ChatGLM3-32B-SFT} & \begin{CJK}{UTF8}{gbsn}设这个两位数为 \(10a + b\)，其中 \(a\) 是十位数字，\(b\) 是个位数字。根据题意，我们有以下两个条件：1. 十位数字比个位数字大 2，即 \(a = b + 2\) 2. 两位数字的积比这个两位数小 34，即 \(ab < 10a + b - 34\) 现在我们来解这个两个方程。从第一个条件，我们得到 \(a = b + 2\)。将 \(a\) 替换进第二个条件中，我们得到：\(b(b + 2) < 10(b + 2) - 34\) 展开并简化这个不等式：\(b^2 + 2b < 10b + 20 - 34\) \(b^2 + 2b < 7b - 14\) 移项得到：\(b^2 - 5b + 14 < 0\) 这是一个一元二次不等式，我们可以求解这个不等式得到 \(b\) 的范围。因式分解得到：\((b - 1)(b - 14) < 0\) 根据一元二次不等式的性质，我们知道 \(b\) 应该在两个根之间，即：\(1 < b < 14\) 但是由于 \(a = b + 2\)，且 \(a\) 是一个十位数（即 \(1 \leq a \leq 9\)），所以 \(b\) 必须小于 9。因此 \(b\) 的可能取值为 \(2, 3, 4, 5, 6, 7, 8\)。现在我们来检查 \(b\) 的每一个值，并找出对应的 \(a\) 和两位数。当 \(b = 2\) 时，\(a = 0\)，这不满足 \(a\) 为十位数的要求；当 \(b = 3\) 时，\(a = 1\)，这也不满足 \(a\) 为十位数的要求；...当 \(b = 7\) 时，\(a = 5\)，这满足要求，并且 \(a\) 和 \(b\) 的积是 \(35\)，确实比 \(35\) 小 34，满足题目的要求。所以这个两位数是 57。\end{CJK} 
\\ \hdashline
\textbf{ChatGLM3-32B-SFT (Translate)} & Let the two-digit number be \(10a + b\), where \(a\) is the tens digit and \(b\) is the units digit. According to the problem, we have two conditions: 1. The tens digit is 2 greater than the units digit, i.e., \(a = b + 2\) 2. The product of the digits is 34 less than the number itself, i.e., \(ab < 10a + b - 34\) Now let's solve these two equations. From the first condition, we get \(a = b + 2\). Substituting \(a\) into the second condition, we get: \(b(b + 2) < 10(b + 2) - 34\) Expanding and simplifying this inequality gives: \(b^2 + 2b < 10b + 20 - 34\) \(b^2 + 2b < 7b - 14\) Rearranging gives: \(b^2 - 5b + 14 < 0\) This is a quadratic inequality, which we can solve to find the range for \(b\). Factoring gives: \((b - 1)(b - 14) < 0\) According to the properties of quadratic inequalities, we know \(b\) should be between the two roots, i.e., \(1 < b < 14\). However, since \(a = b + 2\) and \(a\) is a tens digit (\(1 \leq a \leq 9\)), \(b\) must be less than 9. Thus, the possible values for \(b\) are \(2, 3, 4, 5, 6, 7, 8\). Now we check each value of \(b\) to find the corresponding \(a\) and the two-digit number. When \(b = 2\), \(a = 0\), which does not meet the requirement for \(a\) to be a tens digit; when \(b = 3\), \(a = 1\), which also does not meet the requirement; ... when \(b = 7\), \(a = 5\), which meets the requirement, and the product of \(a\) and \(b\) is \(35\), indeed less than \(35\) by 34, meeting the problem's requirement. Therefore, the two-digit number is 57. \\
\hline
\textbf{ChatGLM3-32B-Math} & \begin{CJK}{UTF8}{gbsn}设这个两位数为 \(10a + b\)，其中 \(a\) 是十位数字，\(b\) 是个位数字。根据题目条件，我们有以下两个方程：1. 十位数字比个位数字大 2，即 \(a = b + 2\)。2. 两位数字的积比这个两位数小 34，即 \(ab = (10a + b) - 34\)。我们可以用第一个方程代入第二个方程来解这个问题。将 \(a = b + 2\) 代入第二个方程，得到：\(b(b + 2) = 10(b + 2) + b - 34\) 展开并简化：\(b^2 + 2b = 10b + 20 + b - 34\) \(b^2 + 2b = 11b - 14\) 移项得到一个二次方程：\(b^2 - 9b + 14 = 0\) 这是一个二次方程，我们可以通过分解因式来解它：\((b - 7)(b - 2) = 0\) 所以，\(b\) 的值可以是 7 或者 2。如果 \(b = 7\)，则 \(a = b + 2 = 9\)，所以这个两位数是 97。如果 \(b = 2\)，则 \(a = b + 2 = 4\)，所以这个两位数是 42。因此，这个两位数是 42 或 97。\end{CJK} 
\\ \hdashline
\textbf{ChatGLM3-32B-Math (Translate)} & Let the two-digit number be \(10a + b\), where \(a\) is the tens digit and \(b\) is the units digit. According to the problem conditions, we have two equations: 1. The tens digit is 2 greater than the units digit, i.e., \(a = b + 2\). 2. The product of the digits is 34 less than the number itself, i.e., \(ab = (10a + b) - 34\). We can solve this problem by substituting the first equation into the second. Doing so gives: \(b(b + 2) = 10(b + 2) + b - 34\) Expanding and simplifying yields: \(b^2 + 2b = 10b + 20 + b - 34\) \(b^2 + 2b = 11b - 14\) Rearranging into a quadratic equation gives: \(b^2 - 9b + 14 = 0\) This quadratic equation can be solved by factoring: \((b - 7)(b - 2) = 0\) Thus, \(b\) can be either 7 or 2. If \(b = 7\), then \(a = b + 2 = 9\), making the two-digit number 97. If \(b = 2\), then \(a = b + 2 = 4\), making the two-digit number 42. Therefore, the two-digit number could be either 42 or 97. \\
\hline
\end{tabular}
}
}
\end{table*}

\begin{table*}[]
\footnotesize
\setlength{\tabcolsep}{15pt}
\caption{ChatGLM3-32B-Math Example (2)}
\label{Math Example (2)}
\centering
	\renewcommand\arraystretch{2}
 \noindent\makebox[\textwidth]{
\resizebox{\textwidth}{!}{

\begin{tabular}{p{0.1\textwidth}p{0.9\textwidth}}

\hline
\textbf{Question} & \begin{CJK}{UTF8}{gbsn}两个相互独立的事件A，B都不发生的概率为1/9，A发生而B不发生的概率与A不发生B发生的概率相等，则P(A)等于？\end{CJK} 
\\ \hdashline
\textbf{Question (Translate)} & Given two independent events A and B, where the probability of neither A nor B occurring is 1/9, and the probability of A occurring without B equals the probability of B occurring without A, what is P(A)? \\
\hline
\textbf{Reference} & \begin{CJK}{UTF8}{gbsn}由题意的：$P(A\bar{B} )=P(B\bar{A} )$,即P(A)(1-P(B))=P(B)(1-P(A)) 所以P(A)=P(B) 又$P(\bar{A}\bar{B})=\frac{1}{9}$  所以$P(\bar{A})=\frac{1}{3}$  所以$P(A)=\frac{2}{3}$\end{CJK} 
\\ \hdashline
\textbf{Reference (Translate)} & From the problem statement: P(A and not B) = P(B and not A), i.e., P(A)(1-P(B)) = P(B)(1-P(A)). Thus, P(A) = P(B). Also, P(neither A nor B) = 1/9. Therefore, P(not A) = 1/3, leading to P(A) = 2/3. \\
\hline
\textbf{GPT-4-0613} & \begin{CJK}{UTF8}{gbsn}设A发生的概率为p，B发生的概率为q，则A，B都不发生的概率为(1-p)(1-q)=1/9，A发生而B不发生的概率为p(1-q)，A不发生B发生的概率为(1-p)q。由题意知，p(1-q)=(1-p)q，解得p=q。  将p=q代入(1-p)(1-q)=1/9，得$(1-p)^2=1/9$，解得p=2/3或p=-1/3。由于概率不能为负，所以p=2/3，即P(A)=2/3。\end{CJK} 
\\ \hdashline
\textbf{GPT-4-0613 (Translate)} & Let the probability of A occurring be p, and the probability of B occurring be q. Then, the probability of neither A nor B occurring is (1-p)(1-q) = 1/9. The probability of A occurring without B is p(1-q), and the probability of B occurring without A is (1-p)q. From the problem, p(1-q) = (1-p)q, leading to p = q. Substituting p = q into (1-p)(1-q) = 1/9 gives $(1-p)^2 = 1/9$, resulting in p = 2/3 or p = -1/3. Since probability cannot be negative, p = 2/3, i.e., P(A) = 2/3. \\
\hline
\textbf{ChatGLM3-32B-SFT} & \begin{CJK}{UTF8}{gbsn}设事件A发生的概率为P(A)，事件B发生的概率为P(B)。已知A和B相互独立，那么事件A不发生而B发生的概率为P(B)(1-P(A))，事件A发生而B不发生的概率为P(A)(1-P(B))。  根据题目条件，A发生而B不发生的概率与A不发生B发生的概率相等，即：  P(A)(1-P(B)) = P(B)(1-P(A))  由于A和B相互独立，我们可以将上式简化为：  P(A) - P(A)P(B) = P(B) - P(B)P(A)  由于A和B的概率都是非负的，上式意味着P(A) = P(B)。但独立事件的概率相等并不一定意味着A和B是同一个事件，因为它们可以有相同的或不同的概率但独立性保持不变。  现在，我们知道了两个事件独立，且都不发生的概率为1/9。这意味着：  P(A) * P(B) = 1/9  由于P(A) = P(B)，我们可以将这个等式写成：  $P(A)^2 = 1/9$  因为P(A)是概率，所以它必须在0和1之间。唯一满足这个等式的概率值是：  $P(A) = √(1/9) = 1/3$  因此，P(A) = 1/3。既然P(A) = P(B)，那么P(B)也等于1/3。\end{CJK} 
\\
\hline
\textbf{ChatGLM3-32B-Math} & \begin{CJK}{UTF8}{gbsn}设事件A发生的概率为P(A)，事件B发生的概率为P(B)。根据题意，我们可以列出以下关系：  1. 事件A和事件B都不发生的概率为1/9，即： \[ P(\text{A不发生且B不发生}) = P(\text{A不发生}) \times P(\text{B不发生}) = (1 - P(A)) \times (1 - P(B)) = \frac{1}{9} \]  2. 事件A发生而事件B不发生的概率与事件A不发生而事件B发生的概率相等，即： \[ P(\text{A发生且B不发生}) = P(\text{A不发生且B发生}) \]  根据概率的性质，我们有： \[ P(\text{A发生且B不发生}) = P(A) \times (1 - P(B)) \] \[ P(\text{A不发生且B发生}) = (1 - P(A)) \times P(B) \]  由于这两个概率相等，我们可以得到： \[ P(A) \times (1 - P(B)) = (1 - P(A)) \times P(B) \]  现在我们有两个方程：  1. \( (1 - P(A)) \times (1 - P(B)) = \frac{1}{9} \) 2. \( P(A) \times (1 - P(B)) = (1 - P(A)) \times P(B) \)  我们可以通过这两个方程来解出P(A)和P(B)的值。  首先，从第二个方程中我们可以得到： \[ P(A) - P(A) \times P(B) = P(B) - P(A) \times P(B) \] \[ P(A) = P(B) \]  这意味着事件A和事件B发生的概率相等。  现在，我们用这个结论来解第一个方程： \[ (1 - P(A)) \times (1 - P(A)) = \frac{1}{9} \] \[ (1 - P(A))^2 = \frac{1}{9} \]  取平方根得到： \[ 1 - P(A) = \frac{1}{3} \quad \text{或} \quad 1 - P(A) = -\frac{1}{3} \]  由于概率值不能为负，我们只取正值： \[ 1 - P(A) = \frac{1}{3} \] \[ P(A) = \frac{2}{3} \]  所以，事件A发生的概率P(A)等于2/3。\end{CJK} \\
\hline
\end{tabular}
}
}
\end{table*}

\begin{table*}[]
\caption{ChatGLM3-32B-Math Example (3)}
\label{Math Example (3)}
\centering
	\renewcommand\arraystretch{2}
\resizebox{\textwidth}{!}{

\begin{tabular}{p{0.13\textwidth}p{0.87\textwidth}}
\hline
\textbf{Question} & The first term of an arithmetic sequence is 18. The sum of the first six terms of the sequence is equal to the sum of the first seven terms.Show that the sum of the first 13 terms of the sequence is 0 and calculate the 13th term of the sequence. \\
\hline
\textbf{Reference} & If the sum of the first six terms of a sequence is equal to the sum of the first seven terms then the seventh term must be 0.  Let d be the common difference of the sequence. Now 18 + 6d = 0 and so d = -3. a13 = 18 + 12*(-3) = -18 \\

\hline
\textbf{ChatGLM3-32B-Math} & Let's denote the common difference of the arithmetic sequence as \( d \). The formula for the \( n \)-th term of an arithmetic sequence is \( a_n = a_1 + (n - 1)d \), where \( a_1 \) is the first term and \( n \) is the term number.

Given that the first term \( a_1 = 18 \), the formula for the \( n \)-th term becomes \( a_n = 18 + (n - 1)d \).

The sum \( S_n \) of the first \( n \) terms of an arithmetic sequence is given by the formula \( S_n = \frac{n}{2} [2a_1 + (n - 1)d] \).

According to the problem, the sum of the first six terms (\( S_6 \)) is equal to the sum of the first seven terms (\( S_7 \)). We can write this relationship as:

\( S_6 = S_7 \)

\( \frac{6}{2} [2 \cdot 18 + (6 - 1)d] = \frac{7}{2} [2 \cdot 18 + (7 - 1)d] \)

\( 3 [36 + 5d] = \frac{7}{2} [36 + 6d] \)

\( 108 + 15d = \frac{7}{2} \cdot 36 + \frac{7}{2} \cdot 6d \)

\( 108 + 15d = 126 + 21d \)

Now, we solve for \( d \):

\( 15d - 21d = 126 - 108 \)

\( -6d = 18 \)

\( d = -3 \)

Now that we have the common difference \( d = -3 \), we can find the 13th term \( a_{13} \):

\( a_{13} = 18 + (13 - 1) \cdot (-3) \)

\( a_{13} = 18 - 12 \cdot 3 \)

\( a_{13} = 18 - 36 \)

\( a_{13} = -18 \)

Now, let's find the sum of the first 13 terms \( S_{13} \) using the formula:

\( S_{13} = \frac{13}{2} [2 \cdot 18 + (13 - 1) \cdot (-3)] \)

\( S_{13} = \frac{13}{2} [36 - 12 \cdot 3] \)

\( S_{13} = \frac{13}{2} [36 - 36] \)

\( S_{13} = \frac{13}{2} \cdot 0 \)

\( S_{13} = 0 \)

So, the sum of the first 13 terms of the sequence is indeed 0, and we've also calculated that the 13th term of the sequence is \( -18 \). \\
\hline

\end{tabular}

}
\end{table*}

\begin{table*}[]
\caption{ChatGLM3-32B-Math Error Case (1)}
\label{Math Error(1)}
\centering
	\renewcommand\arraystretch{2}
\resizebox{\textwidth}{!}{

\begin{tabular}{|p{0.13\textwidth}|p{0.87\textwidth}|}
\hline
\textbf{Question} & The matrices \[\begin{pmatrix} 2 & a \\ -3 & -1 \end{pmatrix} \quad \text{and} \quad \begin{pmatrix} -\frac{1}{16} & b \\ \frac{3}{16} & \frac{1}{8} \end{pmatrix}\]are inverses.  Enter the ordered pair $(a,b).$ \\
\hline
\textbf{Reference} & $(6,-\frac{3}{8})$ \\
\hline
\textbf{ChatGLM3-32B-Math} & Two matrices are inverses if their product is the identity matrix. The identity matrix for a 2x2 matrix is
\[\begin{pmatrix} 1 & 0 \\ 0 & 1 \end{pmatrix}.\]

So, we need to find the values of $a$ and $b$ such that
\[\begin{pmatrix} 2 & a \\ -3 & -1 \end{pmatrix} \begin{pmatrix} -\frac{1}{16} & b \\ \frac{3}{16} & \frac{1}{8} \end{pmatrix} = \begin{pmatrix} 1 & 0 \\ 0 & 1 \end{pmatrix}.\]

We can calculate the product as follows:
\[\begin{pmatrix} 2 & a \\ -3 & -1 \end{pmatrix} \begin{pmatrix} -\frac{1}{16} & b \\ \frac{3}{16} & \frac{1}{8} \end{pmatrix} = \begin{pmatrix} 2(-\frac{1}{16}) + a(\frac{3}{16}) & 2b + a(\frac{1}{8}) \\ -3(-\frac{1}{16}) - 1(\frac{3}{16}) & -3b - 1(\frac{1}{8}) \end{pmatrix}.\]

This gives us the equations
$2(-\frac{1}{16}) + a(\frac{3}{16}) = 1$, 

$2b + a(\frac{1}{8}) = 0$, 

$-3(-\frac{1}{16}) - 1(\frac{3}{16}) = 0$

$-3b - 1(\frac{1}{8}) = 1$

Simplifying, we get
$-\frac{1}{8} + \frac{3a}{16} = 1$, 

$2b + \frac{a}{8} = 0$, 

$\frac{3}{16} - \frac{3}{16} = 0$,

$-3b - \frac{1}{8} = 1$.

From the third equation, we have $\frac{a}{8} = 2b$.  Multiplying both sides by 8, we get $a = 16b$.

Substituting into the first equation, we get \[-\frac{1}{8} + \frac{3(16b)}{16} = 1,\]which simplifies to $3b = \frac{5}{8}$.  Therefore, $b = \frac{5}{24}$.

Substituting into the second equation, we get \[2(\frac{5}{24}) + \frac{16b}{8} = 0,\]which simplifies to $b = -\frac{5}{16}$.  This is consistent with our previous calculation, so $b = \frac{5}{24}$ is the correct value.

Finally, substituting into the fourth equation, we get \[-3(\frac{5}{24}) - \frac{1}{8} = 1,\]which simplifies to $a = \frac{1}{16}$.

Therefore, $(a,b) = \boxed{\left( \frac{1}{16}, \frac{5}{24} \right)}$. 
\\ \bottomrule
\end{tabular}

}
\end{table*}

\begin{table*}[]
\caption{ChatGLM3-32B-Math Error Case (2)}
\label{Math Error(2)}
\centering
	\renewcommand\arraystretch{2}
\resizebox{\textwidth}{!}{

\begin{tabular}{|p{0.13\textwidth}|p{0.87\textwidth}|}
\hline
\textbf{Question} & n the graph below, each grid line counts as one unit. The line shown below passes through the point $(1001,n)$ (not shown on graph).  Find $n$.  \begin{center}
\begin{tikzpicture}[scale=0.5]
    \draw[step=1cm,gray,very thin] (-10,-10) grid (10,10);
    \draw[thick,->] (-10,0) -- (10,0) node[anchor=north west] {$x$};
    \draw[thick,->] (0,-10) -- (0,10) node[anchor=south east] {$y$};
    \draw[blue,thick,->] (-10,-2.71) -- (10,8.71) node[anchor=south west] {};
\end{tikzpicture}
\end{center}

\textbf{(note: the graph was written in Asymptote code in original question)}

\\
\hline
\textbf{Reference} & 575 \\
\hline
\textbf{ChatGLM3-32B-Math} & 
The line passes through the point $(1001,n)$, which means that when $x = 1001$, $y = n$. We can find the slope of the line by looking at the change in $y$ divided by the change in $x$ between two points on the line.

If we consider the point $(999, 7)$ and the point $(1001, n)$, the change in $x$ is $1001 - 999 = 2$ and the change in $y$ is $n - 7$.

Since the line passes through the point $(1001, n)$, we can say that the slope of the line is $\frac{n - 7}{2}$.

We know that the line also passes through the point $(0, -2.71)$, so we can use this point to find the slope of the line. The change in $x$ is $0 - (-10) = 10$ and the change in $y$ is $-2.71 - (-7) = 4.29$.

Now we can set the two slopes equal to each other and solve for $n$:

\[\frac{n - 7}{2} = \frac{4.29}{10}\]

Cross multiplying gives:  \[10(n - 7) = 2 \cdot 4.29\]

\[10n - 70 = 8.58\]

Adding 70 to both sides gives:

\[10n = 78.58\]

Finally, dividing both sides by 10 gives:  \[n = 7.858\]

But since the grid lines count as one unit, we need to round $n$ to the nearest whole number. $7.858$ is closer to $8$ than it is to $7$, so $n = \boxed{8}$.
 \\ \bottomrule
\end{tabular}

}
\end{table*}

\begin{table*}[]
\caption{ChatGLM3-32B-Math Error Case (3)}
\label{Math Error(3)}
\centering
	\renewcommand\arraystretch{2}
\resizebox{\textwidth}{!}{

\begin{tabular}{|p{0.13\textwidth}|p{0.87\textwidth}|}
\hline
\textbf{Question} & In the year 2018 factory A produced goods worth 500 million forints, while factory B produced goods worth 400 million forints. According to their long-term plans, factory A will increase production by 5\% each year, while factory B will increase production by 6\% per year.

The above plans were published on an economy website. A debate started in the comment section of the article: one user claimed that the difference between the volumes of production between the two factories would get smaller and smaller in the next couple of years.

Calculate the missing values of the table below and show that the above claim is not true.

| Year | Production volume of factory A (million Ft) | Production volume of factory B (million Ft) |

|------|--------------------------------------------|---------------------------------------------|

| 2018 | 500                                        | 400                                         |

| 2019 |                                            |                                             |

| 2020 |                                            |                                             |

| 2021 |                                            |                                             |
\\
\hline
\textbf{Reference} & | Year | Production volume of factory A (million Ft) | Production volume of factory B (million Ft) |

|------|--------------------------------------------|---------------------------------------------|

| 2018 | 500                                        | 400                                         |

| 2019 | 525                         | 424                          |

| 2020 | 551.3                    | 449.4                       |

| 2021 | 578.8                  | 476.4                   | 

The difference between the production values in the 
given years: 

525 – 424 = 101, 

551.3 – 449.4 = 101.9 and 

578.8 – 476.4 = 102.4 (million forints).

As the difference between the production values is 
increasing (in the period in question), the statement 
is, in fact, false. 
\\
\hline
\textbf{ChatGLM3-32B-Math} & To calculate the production volumes for the years 2019, 2020, and 2021, we will apply the percentage increases to the production of each factory.

For Factory A:
- 2018 production: 500 million Ft
- 5

For Factory B:
- 2018 production: 400 million Ft
- 6

Let's fill in the table:

| Year | Production volume of factory A (million Ft) | Production volume of factory B (million Ft) |

|------|--------------------------------------------|---------------------------------------------|

| 2018 | 500                                        | 400                                         |

| 2019 | 500 * 1.05 = 525                         | 400 * 1.06 = 424                          |

| 2020 | 525 * 1.05 = 551.25                     | 424 * 1.06 = 448.24                       |

| 2021 | 551.25 * 1.05 = 578.96                  | 448.24 * 1.06 = 475.92                   |

Now, let's analyze the data. The difference in production between the two factories in 2018 was 100 million Ft (500 million - 400 million).

- In 2019, the difference is 21 million Ft (525 million - 424 million).
- In 2020, the difference is 23.31 million Ft (551.25 million - 448.24 million).
- In 2021, the difference is 23.94 million Ft (578.96 million - 475.92 million).

The difference is not getting smaller; in fact, it is increasing slightly each year. This is because the rate of increase for Factory A (5\%) is lower than the rate for Factory B (6\%). The claim made in the debate that the difference would get smaller and smaller is not true based on these figures. \\ \bottomrule
\end{tabular}

}
\end{table*}

Here are a few comparisons between ChatGLM3-32B-Math(ChatGLM3-32B-SFT-2312 + RFT\&DPO) and other models. In the example from Table~\ref{Math Example (1)}, the problem is a math question of Chinese junior high school difficulty. During the solution process by GPT-4-0613, an error occurred in solving the equation. ChatGLM3-32B-SFT-2312 did not correctly understand the question. ChatGLM3-32B-Math correctly listed the equation and accurately solved it using the factorization method.

In the example from Table~\ref{Math Example (2)}, both GPT-4-0613 and ChatGLM3-32B-Math provided the correct answers, but the difference lies in that ChatGLM3-32B-Math offered a very detailed derivation process. We believe that detailed derivation aids in understanding for users and helps prevent errors that may occur during the model's step-skipping.

In the example from Table~\ref{Math Example (3)}, originating from the Hungry Test, ChatGLM3-32B-Math correctly conducted the analysis and provided the solution. In contrast, Qwen-Max, despite being accurate in most processes, made a simplification error in the expressions for S6 and S7, leading to an incorrect result despite precise calculations.


\subsection{Case Study of Errors}

In the first error example shown in Table~\ref{Math Error(1)}, ChatGLM3-32B-Math correctly listed the matrix equations but made a mistake in transferring terms during the complex calculation process. This led to an incorrect solution despite the overall process being correct. 

In the second example shown in Table~\ref{Math Error(2)}, the question includes an image. This question originates from MATH, and although the image is provided in Asymptote code, the language model still struggles to understand the meaning of the image. This demonstrates a deficiency in our model's ability to process images. In fact, within the MATH test set, for questions that include images, our model's accuracy rate is only 23\%, significantly lower than the 40\% accuracy rate across the entire dataset.

In the third example shown in Table~\ref{Math Error(3)}, we can observe certain deficiencies in the computational accuracy of the language model. In the first column of the table, the model correctly calculates the multiplication of integers and a single decimal. Still, in the second column, when calculating 424*1.06, the model incurred an error of approximately 0.2\%. This error led to subsequent calculations being incorrect due to the accumulation of errors.
\section{Prompt Template}

\begin{figure}
    \centering
    \includegraphics[width=\linewidth]{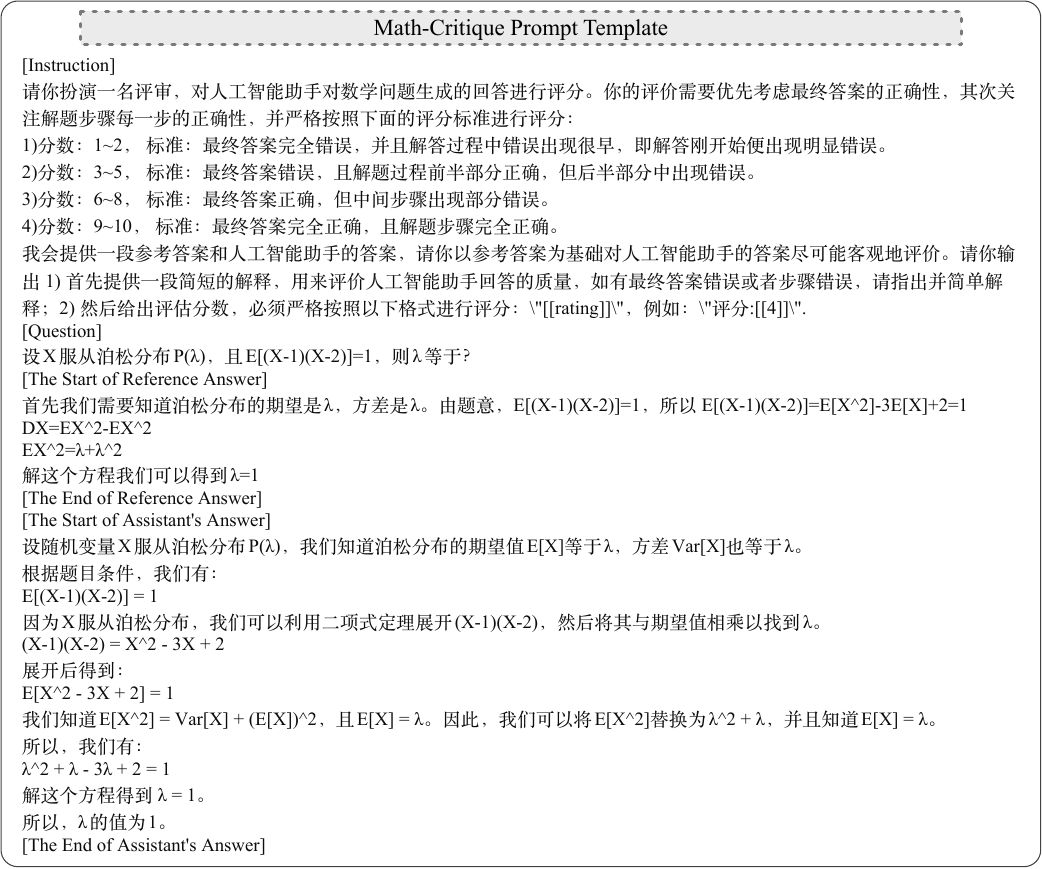}
    \vspace{-5mm}
    \caption{Prompt Template with an example for Math-Critique Model. }
    \label{fig:prompt-example-mc}
\end{figure}

In Figure~\ref{fig:prompt-example-mc}, we present the Chinese version example of the prompts utilized by Math-Critique and the corresponding English translations. In practical application, we employ the Chinese version of the prompts for all inquiries.

\section{Evaluation Settings}
\label{appendix:evaluation}

\subsection{Evaluation Settings for Academic Datasets}
Given our focus on evaluating zero-shot capabilities, traditional methods of answer verification fail to provide accurate assessments. Consequently, we have adapted our evaluation methodology as follows:
\begin{enumerate}
    \item Extract the final sentence of the response (demarcated by a double newline character) or prompt the model with "Therefore, the answer is".
    \item For numeric standard answers, the response undergoes the following recognition process: fractions or decimals in LaTeX format, and regular fractions, decimals, or integers. We compute the numerical result and compare it with the standard answer. A discrepancy less than (1e-6) is deemed correct.
    \item For standard answers that are strings (exclusive to the MATH dataset): identify the content following "xxx is" or an equality sign, or within $boxed\{\}$. The evaluation considers whether the normalized strings match.
\end{enumerate}


For the English datasets GSM8k and MATH, our principle for selecting results is as follows:
\begin{enumerate}
    \item If the model itself reports results on these two datasets, we choose the higher of either the zero-shot or few-shot results.
    \item If the model does not report its results, we utilize the results reported by OpenCompass, again selecting the higher of the zero-shot or few-shot outcomes. This applies to models such as GPT-4-0613, GPT-3.5-Turbo-0613, Yi-6B-Chat, and Qwen-Chat-7B. Should there be any omissions from the above sources, results will not be reported.
\end{enumerate}

For the Hungarian Exam results, we had annotators score them according to a grading standard, which is referenced from \cite{testing_language_models_on_a_held_out_high_school_national_finals_exam}. It's noted that the reported results actually sum the scores of all questions, making the total score effectively 117. To align with the reported results, we adopted this scoring method as well.

With reproducibility in mind, all our results were obtained using a sampling temperature of 0 and setting the max-seq-length to 4096.

\subsection{Evaluation Settings for 2023 Hungarian national high school finals in mathematics}

For the Hungarian national high school finals in mathematics, we submit the model's answers to annotators for marking. For results of models not listed in ~\cite{testing_language_models_on_a_held_out_high_school_national_finals_exam}, we score them based on the answers provided in~\cite{testing_language_models_on_a_held_out_high_school_national_finals_exam} according to the scoring points. We sum the scores of all questions to present a total score. All annotations are carried out by two annotators; in case of inconsistency, a third annotator decides.

Considering the general situation of multiple models, we do not restrict the language used by the language models to answer the questions. Any language used to correctly answer is considered correct. Additionally, since most questions do not restrict the form of the answer, we stipulate that answers are deemed correct as long as they retain more than one decimal place accurately or are provided in fraction form.
\section{Additional Results}

\subsection{Subcategory Results of MathUserEval}
\begin{table*}[t]
\caption{Math-User-Result Result, GPT-4-1106-Preview-rated. \textmd{All results were scored by GPT-4-1106-Preview, with the scoring method consistent with AlignBench. All Overall scores were calculated using the macro-average.}}
\label{table:matheval-GPT4-rated}
\noindent\makebox[\textwidth]{
\resizebox{\textwidth}{!}{

\begin{tabular}{@{}lccccccccccc@{}}
\toprule
\multicolumn{1}{c}{\multirow{2}{*}{Model}} & \multicolumn{1}{c}{\multirow{2}{*}{Overall}} & \multicolumn{5}{c}{Elementary}                                                                                                                              & \multicolumn{5}{c}{Advanced}                                                                                                                                       \\ \cmidrule(l){3-7} \cmidrule(l){8-12} 
\multicolumn{1}{c}{}                       & \multicolumn{1}{c}{}                         & \multicolumn{1}{c}{Avg} & \multicolumn{1}{c}{algebra} & \multicolumn{1}{c}{calculate} & \multicolumn{1}{c}{geo.} & \multicolumn{1}{c}{tri.} & \multicolumn{1}{c}{Avg} & \multicolumn{1}{c}{calculus} & \multicolumn{1}{c}{discrete} & \multicolumn{1}{c}{linear.} & \multicolumn{1}{c}{Prob.}      \\ \midrule
GPT-4-0125-Preview~\cite{openai2023gpt4}                          & \textbf{5.79}                                 & \textbf{5.26}               & \textbf{5.04}               & \textbf{7.63}                 & \textbf{3.98}                & 4.59                              & 6.71                        & 7.26                         & 6.62                         & \textbf{5.48}                       & 7.72                                                        \\
GPT-4-1106-Preview~\cite{openai2023gpt4}                          & 5.73                                          & 5.07                        & 4.96                        & 7.00                             & 3.78                         & 4.71                              & \textbf{6.81}               & \textbf{7.39}                & \textbf{6.96}                & 5.29                                & \textbf{7.91}                                                \\
GLM-4                                        & 5.11                                          & 4.86                        & 4.47                        & 6.56                          & 3.95                         & \textbf{4.74}                     & 5.43                        & 6.00                          & 5.67                         & 4.26                                & 6.02                                                \\
\textbf{ChatGLM3-32B-SFT-2312 + RFT\&DPO}                         & 4.23                                          & 4.01                        & 3.88                        & 5.41                          & 2.90                         & 3.99                              & 4.59                        & 5.22                         & 4.76                         & 3.38                                & 5.20                                                        \\
GPT-4-0613~\cite{openai2023gpt4}                                  & 4.14                                          & 3.34                        & 2.88                        & 4.76                          & 3.17                         & 2.78                              & 5.33                        & 5.57                         & 5.49                         & 4.26                                & 6.22                                                        \\
\textbf{ChatGLM3-32B-SFT-2312 + RFT}                        & 4.01                                          & 3.86                        & 3.84                        & 5.37                          & 2.57                         & 3.77                              & 4.26                        & 4.72                         & 4.69                         & 2.98                                & 4.89                                                   \\
Qwen-72B-Chat~\cite{bai2023qwen}                               & 3.87                                          & 3.99                        & 3.96                        & 4.81                          & 3.83                         & 3.34                              & 3.67                        & 4.54                         & 3.71                         & 2.84                                & 3.65                                            \\
GPT-3.5-Turbo-0613~\cite{openai2023gpt4}                          & 3.42                                          & 3.04                        & 2.81                        & 4.07                          & 2.23                         & 3.26                              & 4.07                        & 4.83                         & 4.38                         & 3.26                                & 3.91                                    \\
\textbf{ChatGLM3-32B-SFT-2312}                    & 3.39                                          & 3.35                        & 3.35                        & 4.51                          & 2.51                         & 3.11                              & 3.44                        & 4.04                         & 4.38                         & 2.41                                & 3.13                                       \\
Claude-2~\cite{Claude}                                    & 3.29                                          & 2.63                        & 2.35                        & 3.63                          & 2.20                          & 2.53                              & 4.35                        & 4.56                         & 4.53                         & 3.29                                & 5.28                                \\
DeepSeek-Chat-67B~\cite{deepseekai2024deepseek}                          & 3.24                                          & 2.76                        & 2.21                        & 4.73                          & 2.12                         & 2.30                               & 3.84                        & 4.41                         & 4.82                         & 2.79                                & 3.52                                 \\
Yi-34B-Chat~\cite{Yi}                                 & 2.64                                          & 2.49                        & 2.04                        & 3.61                          & 2.25                         & 2.27                              & 2.87                        & 2.80                          & 3.47                         & 2.03                                & 3.41                                          \\ \bottomrule
\end{tabular}
}}
\end{table*}
In Table~\ref{table:matheval-GPT4-rated}, we display the results for all subsets of MathUserEval. The reported results were evaluated by GPT-4-1106-Preview, with the evaluation method consistent with AlignBench. It is noted that GPT-4-0125-Preview and GPT-4-1106-Preview still occupy the leading positions. Except for Probability, the GLM4 model's total score and individual scores surpassed GPT-4-0613. Our GLM-Math-32B w/ DPO model performed exceptionally well in the Elementary category, exceeding GPT-4-0613, but a significant gap remains in Advanced mathematics. Our Self-Critique training method showed significant progress in MathUserEval, with an overall improvement of 24\%.

\subsection{Subcategory Results of Alignbench~\cite{liu2023alignbench}}

\begin{threeparttable}[]
\begin{tabular}{@{}lccccccc@{}}
\toprule
\multirow{2}{*}{Model}        & \multicolumn{7}{c}{Language}              \\ \cmidrule(l){2-8} 
                              & \textbf{Avg.}    & \textbf{Fund.}      & \textbf{Chi.}    & \textbf{Open.}  & \textbf{Writ.}    & \textbf{Role.}     & \textbf{Pro.} \\ \midrule
GPT-4-1106-Preview~\cite{openai2023gpt4}            & 8.29 & 7.99 & 7.33 & 8.61 & 8.67 & 8.47 & 8.65 \\
\textbf{ChatGLM3-32B-SFT-2312 + RFT\&DPO}  & 7.80 & 7.14 & 6.90 & 8.37 & 8.41 & 8.09 & 7.90 \\
GPT-4-0613~\cite{openai2023gpt4}                    & 7.59 & 7.81 & 6.93 & 7.42 & 7.93 & 7.51 & 7.94 \\
\textbf{ChatGLM3-32B-SFT-2312 + RFT} & 7.43 & 6.37 & 6.95 & 8.03 & 7.71 & 7.97 & 7.54 \\
\textbf{ChatGLM3-32B-SFT-2312}              & 7.38 & 6.84 & 7.02 & 8.08 & 7.37 & 7.70 & 7.27 \\
Qwen-72B-Chat~\cite{bai2023qwen}                 & 7.29 & 6.63 & 7.31 & 7.24 & 7.29 & 7.59 & 7.71 \\
DeepSeek-67B-Chat~\cite{deepseekai2024deepseek}             & 7.11 & 7.12 & 6.52 & 7.58 & 7.20 & 6.91 & 7.37 \\
GPT-3.5-Turbo-0613~\cite{openai2023gpt4}            & 6.82 & 6.71 & 5.81 & 7.29 & 7.03 & 7.28 & 6.77 \\
Claude-2~\cite{Claude}                      & 6.78 & 6.87 & 6.24 & 7.08 & 6.36 & 6.85 & 7.31 \\
Yi-34B-Chat~\cite{Yi}                   & 6.18 & 4.32 & 6.05 & 7.37 & 6.00 & 6.30 & 7.06 \\ \bottomrule
\end{tabular}
\caption{Results of Alignbench~\cite{liu2023alignbench}, Language Part. }
\label{table:alignbench-details}
\begin{tablenotes}
\item [1] The ChatGLM3-32B-SFT-2312 is a newer version of the ChatGLM series and not identical to the model discussed in ~\cite{hou2024chatglmrlhf}, despite sharing the same model size.
\end{tablenotes}
\end{threeparttable}


Table~\ref{table:alignbench-details} reports detailed results from the language capability subsection of AlignBench. Within this, we present the scores of our four models and have tested the results for Qwen-72B-Chat~\cite{bai2023qwen}, Claude-2~\cite{Claude}, and Yi-34B-Chat~\cite{Yi}. Additional results are derived from the AlignBench paper, and the results for DeepSeek are taken from its report~\cite{deepseekai2024deepseek}.

\end{document}